\documentclass{article} 
\usepackage[preprint]{colm2025_conference} 

\usepackage{microtype}
\usepackage{hyperref}
\usepackage{url}
\usepackage{booktabs}

\usepackage{lineno}

\usepackage{amsmath}
\usepackage{subcaption}
\usepackage{makecell}
\usepackage{arydshln}
\usepackage{pifont}
\usepackage{adjustbox}
\usepackage{booktabs}
\usepackage{multirow}
\usepackage{colortbl}
\usepackage{amssymb}
\usepackage{upquote}
\usepackage{xcolor}
\usepackage{graphicx}
\usepackage{wrapfig}

\definecolor{darkblue}{rgb}{0, 0, 0.5}
\hypersetup{colorlinks=true, citecolor=darkblue, linkcolor=darkblue, urlcolor=darkblue}

\definecolor{fig1Pink}{RGB}{255,205,205}
\definecolor{fig1Blue}{RGB}{206,227,255}
\definecolor{tab1Blue}{RGB}{193,206,246}
\definecolor{tab1Blue2}{RGB}{0,68,158}

\title{\raisebox{-0.1cm}{\includegraphics[width=1.2em]{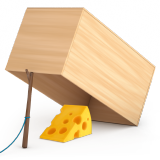}} The Comparative Trap: Pairwise Comparisons Amplifies\\Biased Preferences of LLM Evaluators}

\author{
Hawon Jeong\thanks{\hspace{0.2cm} Equal contribution}\hspace{0.2cm}$^1$ \hspace{0.3cm} 
ChaeHun Park\footnotemark[1]\hspace{0.2cm}$^1$ \hspace{0.3cm} 
Jimin Hong$^1$$^2$ \hspace{0.3cm} 
Hojoon Lee$^1$ \hspace{0.3cm} 
Jaegul Choo$^1$ \\
$^1$ KAIST AI \hspace{0.3cm} $^2$ KRAFTON\\
\hspace{0cm}\texttt{\{hawon,ddehun,jimmyh,joonleesky,jchoo\}@kaist.ac.kr} \\
 }

\begin{document}

\ifcolmsubmission
\linenumbers
\fi

\maketitle

\begin{abstract}
As large language models (LLMs) are increasingly used as evaluators for natural language generation tasks, ensuring unbiased assessments is essential. However, LLM evaluators often display biased preferences, such as favoring verbosity and authoritative tones.
Our empirical analysis reveals that these biases are exacerbated in pairwise evaluation, where LLMs directly compare two outputs and easily prioritize superficial attributes.
In contrast, pointwise evaluation, which assesses outputs independently, is less susceptible to such bias because each output is judged in isolation.
To address the limitations of the pairwise evaluation, we introduce a novel evaluation method, \textsc{PRePair}, which integrates pointwise reasoning within a pairwise framework. \textsc{PRePair} effectively alleviates biased preference, improving performance on the adversarial benchmark (LLMBar) while outperforming pointwise evaluation on the standard benchmark (MT-Bench).

\end{abstract}

\section{Introduction}
\label{sec:introduction}

Automated evaluation systems are crucial for scaling natural language generation (NLG) research, offering a more efficient and cost-effective alternative to human evaluation. Recently, large language models (LLMs) have been increasingly adopted as automatic evaluators, demonstrating promising capabilities in assessing various aspects of text quality such as coherence, relevance, and fluency~\citep{chiang2023can, liu2023gpteval}. 


Despite their impressive performance, LLM-based evaluators often display systematic biases in their preferences—for instance, favoring verbose responses~\citep{Wang2023LargeLM}, exhibiting egocentric judgment~\citep{zheng2024judging}, or preferring authoritative tones~\citep{chen2024humans}.
Recent studies have begun to investigate these biases by probing the evaluator’s ability to follow explicit evaluation instructions in controlled settings~\citep{zeng2024llmbar, lambert2024rewardbench}. 
In these evaluations, one output strictly follows the instructions, while the other is adversarially crafted to exploit evaluator biases—appearing superficially favorable yet failing to comply with the instructions. These studies reveal that LLM evaluators not only exhibit biased preferences but also struggle to identify these deceptive outputs effectively.

Building on these observations, we first explore how different evaluation setups influence the biased preferences in LLM evaluators. Specifically, we compare two common setups: \textit{pairwise}, where the evaluator compares two outputs simultaneously and selects the better one, and \textit{pointwise}, where each output is evaluated independently and then selects the one with a higher score. Our experiments demonstrate that the pairwise setup amplifies the biases of the evaluator, performing significantly worse than the pointwise setup on adversarial examples (see Fig.~\ref{fig:analysis_comparison} Right). 
Further analysis reveals that pairwise evaluators, while recognizing weaknesses in incorrect outputs, often reach incorrect conclusions.

We hypothesize that the increased bias in pairwise evaluations stems from the structural differences between the pairwise and pointwise setups. To illustrate, consider a job interview where the interviewer must choose between Alice and Bob for a research engineer position. Although the goal is to select the candidate more suited for the role, the interviewer might favor Alice simply if she has more fluent speaking skills
despite Bob possibly having stronger technical skills. Psychological studies support this phenomenon: when directly comparing options, humans often focus on superficial features rather than thoroughly evaluating each option~\citep{wright1974harassed, payne1976task, olshavsky1979task}. 
Similarly, in a pairwise setup, LLMs may concentrate on easily exploitable, superficial attributes, causing adversarial outputs to appear favorable even when they fail to follow the instructions.

In contrast, pointwise evaluation assesses each output independently, which allows the model to focus solely on how well the output adheres to the instructions, without being interrupted by comparative attributes. This isolation reduces susceptibility to biases and leads to a more accurate assessment of task compliance. However, pointwise evaluation is not a silver bullet for all scenarios; Suppose Alice and Bob are equally fluent speakers. 
In this case, evaluating both candidates simultaneously can provide additional context about who is ultimately better suited for the role. 
In such cases, the pairwise approach benefits from leveraging contextual information for non-adversarial examples (See Fig. \ref{fig:analysis_comparison} Left).

Motivated by our findings, we propose a simple yet effective method for reliable pairwise evaluation, namely \textsc{PRePair}, which stands for \textsc{\textbf{P}}ointwise \textsc{\textbf{Re}}asoning within a \textsc{\textbf{Pair}}wise evaluation framework (Fig.~\ref{fig:method}).
The key idea is to first extract unbiased insights from each output using pointwise reasoning. Then, we combine these individual explanations into a final decision process for pairwise evaluation, leveraging the strength of both pointwise and pairwise approaches.
Experimental results across various models demonstrate that \textsc{PRePair} improves the performance of pairwise evaluators on adversarial datasets. 
Specifically, \textsc{PRePair} improves average accuracy by 24.48\% across six LLMs on the adversarial dataset.
Moreover, it outperforms the pointwise approach on normal datasets, demonstrating its versatility and effectiveness. 

Our contributions can be summarized as follows:  
(1) We analyze how pointwise and pairwise evaluation setups affect LLM assessment in adversarial scenarios.
(2) We identify biases in pairwise evaluation that favor superficial attributes, making adversarial responses more likely to be preferred.   
(3) We propose a hybrid evaluation framework by integrating pointwise reasoning into pairwise evaluation.  

\section{Related Work}
\label{sec:related_work}


\noindent{\textbf{LLMs-based Evaluation}} The evaluation of NLG systems has traditionally relied on human annotation and automatic metrics such as BLEU~\citep{papineni-etal-2002-bleu}, ROUGE~\citep{lin-2004-rouge}, and METEOR~\citep{banerjee-lavie-2005-meteor}. However, these metrics often fail to capture semantic adequacy and coherence~\citep{liu-etal-2016-evaluate,sai2022survey}. To address these limitations, large language models (LLMs) have emerged as evaluators due to their strong instruction-following capabilities and generalization across diverse tasks~\citep{kocmi2023nmt-qe-llm, lin2023llm, min-etal-2023-factscore, chan2023chateval, wang2024self}.

Among various evaluation strategies, pairwise comparison has been increasingly utilized, where an LLM ranks or selects the better response between two candidates~\citep{chiang2024chatbot, zeng2024llmbar, Wang2023LargeLM, zhang2023wider}. This method aligns with human evaluation practices and has been shown to correlate well with human judgments~\citep{zheng2024judging}. However, LLM evaluators exhibit systematic biases, such as favoring verbose responses~\citep{li2023prd, zheng2024judging} and overrating their own generations~\citep{li2023prd, koo2023benchmarking}. Moreover, their rankings can be inconsistent, as slight prompt variations may lead to different outcomes~\citep{liu2023gpteval, zhu2024promptbench}. Recent studies suggest that incorporating explicit evaluation criteria and general guidelines in prompts can improve their discernment~\citep{zeng2024llmbar}. Building on these insights, we examine how different evaluation setups influence LLM-based assessment, particularly in adversarial scenarios where biases may be exacerbated.

\noindent{\textbf{Human Behavior with Complex Tasks}} 
Psychological research has demonstrated that decision-makers often simplify their process by focusing on limited information when handling complex tasks~\citep{payne1976task, olshavsky1979task, einhorn1981behavioral}. 
Moreover, \citet{wright1974harassed} has indicated that cognitive overload from environmental distractions reduces decision quality, causing people to overlook important evidence or rely on irrelevant information.
These findings parallel our intuitions in adversarial evaluation: assessing multiple outputs at once encourages the model to prioritize superficial aspects, similar to humans favoring simpler solutions in complex tasks.

\section{Analyzing LLM Evaluators on Adversarial Evaluation Samples}
\label{sec:analysis}

We now analyze LLM-based evaluators using different datasets and prompting strategies. We introduce meta-evaluation datasets with different characteristics and then compare two representative evaluation approaches on these datasets. Finally, we examine the impacts of intrinsic biases in LLM evaluators by exploring their reasoning process.

\subsection{Meta-Evaluation Datasets}
In our experiments, we evaluate LLM evaluators using datasets with two distinct characteristics: \textbf{Normal} and \textbf{Adversarial}. The normal dataset, \textit{MT-Bench}~~\citep{zheng2024judging}, contains responses generated under standard conditions and reflects real-world evaluation distributions. In contrast, the adversarial dataset, \textit{LLMBar-Adversarial}~~\citep{zeng2024llmbar}, consists of responses that appear high-quality but subtly deviate from the original instruction, testing an evaluator's ability to detect misleading outputs.
Each evaluation sample in both datasets consists of (1) an input instruction, (2) a pair of two model-generated responses, and (3) a human preference label indicating which response is preferred. The datasets contain a total of 1131 evaluation samples from \textit{MT-Bench} and 319 from \textit{LLMBar-Adversarial}. Further details on each dataset are provided below.

\noindent{\textit{{MT-Bench}}~\citep{zheng2024judging} is a meta-evaluation dataset consisting of 80 multi-turn instructions, each paired with model-generated outputs from various language models. For our experiments, we use the first-turn meta-evaluation samples. The dataset includes a total of 1131 human judgments from two groups: expert labelers (885) and the original authors (246). This dataset is widely used to assess evaluation metrics in a setting that reflects real-world model performance.

\noindent{\textit{LLMBar-Adversarial}}~\citep{zeng2024llmbar} is specifically designed to test the ability of evaluators to discern instruction-following responses. The dataset consists of 319 adversarial evaluation samples, where generated responses exhibit high superficial quality but fail to adhere to the original instruction. These samples are categorized into four types: Neighbor Instructions (134), GPT-4 Instructions (92), GPT-4 Unhelpful Outputs (47), and Manual Construction (46). This dataset serves as a benchmark for evaluating whether LLM evaluators can detect subtle deviations from the expected responses. 

\subsection{Setup}
\label{sec:analysis-setup}
\noindent{\textbf{Pointwise and Pairwise Approaches}}
We compare the two popular approaches for pairwise text evaluation: \textit{pairwise} and \textit{pointwise}~\citep{zheng2024judging,liusie2024llm}.
In the pairwise approach, an LLM receives an instruction and two model-generated outputs and then chooses a higher-quality 
output.
In the pointwise approach, an LLM assigns a single score on a Likert scale to each generated output independently. The output with the higher score is considered to be of higher quality. Please refer to Appendix \ref{sec:app:eval_impl_detail} for more details.

\noindent{\textbf{Implementation Details}}
We use the following LLMs as evaluators to compare the approaches above (i.e., pointwise and pairwise) on MT-Bench and LLMBar datasets: \texttt{Llama-3.1-8B-Instruct}~\citep{dubey2024llama3}, \texttt{gpt-3.5-turbo}, \texttt{gpt-4}, \texttt{gpt-4o-mini}, and \texttt{gpt-4o}~\citep{gpt4}.
We utilize the original prompts released by each dataset for evaluation.
For LLMBar, we include general rules, such as prioritizing outputs that accurately follow instructions, as these guidelines generally enhance evaluation quality~\citep{zeng2024llmbar}.
The performance of the evaluator is obtained by comparing its judgments to human judgments within the datasets.
In MT-Bench, since a tie option is allowed, the pairwise evaluator is prompted to choose one of three options. The scalar value generated by the pointwise evaluator using the Likert-scale is directly used for evaluation.\footnote{{Evaluation results of pointwise evaluators with weighted summation on a dataset that does not allow tie-option (i.e., RewardBench-Chat~\citep{lambert2024rewardbench}) are also presented in Section~\ref{sec:reward-chat}.}}
In LLMBar, the human judgments contain only two options without tie. 
Following \citet{liu2023gpteval}, we use a weighted summation with token probabilities for the pointwise evaluator.

\begin{figure*}[t!]
\centering
\vspace{-0.6cm}
\includegraphics[width=\linewidth]{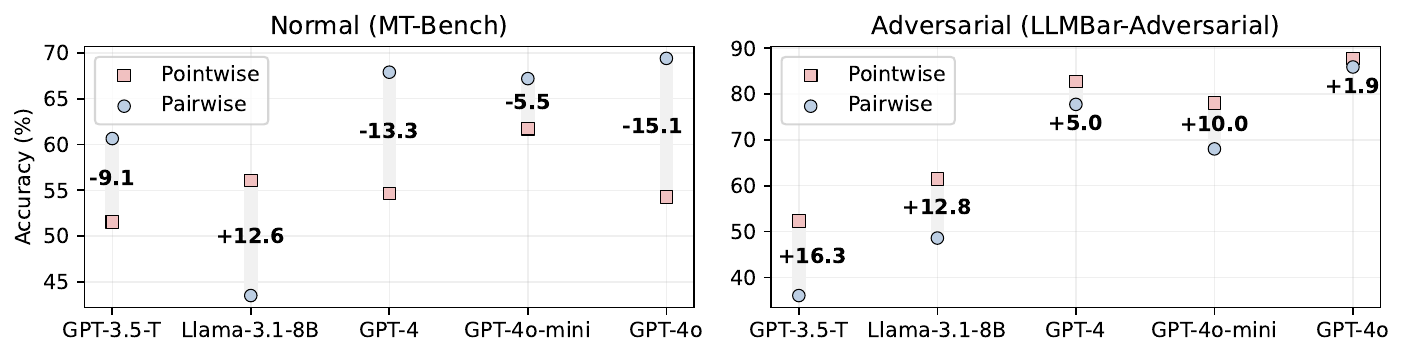}
\vspace{-0.8cm}
\caption{Results of \setlength{\fboxsep}{2pt}
\colorbox{fig1Pink}{pointwise} and \setlength{\fboxsep}{2pt}\colorbox{fig1Blue}{pairwise} approaches in normal~(MT-Bench) and adversarial~(LLMBar-Adversarial) meta-evaluation datasets. The difference values represent the change between pointwise and pairwise approaches for each model under both normal and adversarial conditions.}
\label{fig:analysis_comparison}
\vspace{-0.2cm}
\end{figure*}
\definecolor{customblue}{rgb}{0.812, 0.835, 0.976}
\definecolor{customred}{rgb}{0.961, 0.741, 0.714}

\begin{table}[t!]
\small
\centering
\begin{tabular}{p{\textwidth}}
\bottomrule
\rowcolor[rgb]{0.95,0.95,0.95} \raisebox{-0.2\totalheight}{\textbf{Evaluation Sample}} \\ [0.5ex]
\noalign{\vskip 0.5ex} 
\textbf{Instruction:} Create a new headline that is witty and summarizes the article. A new study has found that eating spinach can increase intelligence. \\ 
\noalign{\vskip 0.5ex} 
\cdashline{1-1}
\noalign{\vskip 0.5ex}
\textbf{Output A}: Spinach: The New Brain Food, or Just Another Leafy Green? (\textcolor{blue}{\textit{\textbf{Human-preferred}}})\\ 
\noalign{\vskip 0.5ex}
\textbf{Output B}: Spinach It To Win: Eating Spinach Can Boost Your Brain Power! \\
\hline 
\rowcolor[rgb]{0.95,0.95,0.95} \raisebox{-0.2\totalheight}{\textbf{Explanations from LLM Evaluator}} \\ [0.5ex]
\noalign{\vskip 0.5ex} 
\textbf{Strengths of Output A}: Clever play on words with "Brain Food" and "Leafy Green". \slash Succinctly summarizes the article topic. \\  
\noalign{\vskip 0.5ex} 
\textbf{Weaknesses of Output A}: Lacks a direct mention of the study findings. \slash \ Could be perceived as slightly less engaging \setlength{\fboxsep}{1pt}\colorbox{red!30}{{compared to Output B}}. \\ 
\noalign{\vskip 0.5ex}
\cdashline{1-1}
\noalign{\vskip 0.5ex}
\textbf{Strengths of Output B}: Engaging with the phrase "Spinach It To Win." \slash \ Clearly states the benefit of eating spinach for brain power.\\ 
\noalign{\vskip 0.5ex}
\textbf{Weaknesses of Output B}: The use of "Win" might be seen as slightly \setlength{\fboxsep}{1pt}
\colorbox{tab1Blue}{exaggerated or sensationalized.} 
\slash \ Could be considered a bit \setlength{\fboxsep}{1pt}\colorbox{tab1Blue}{informal for certain audiences.} \\
\noalign{\vskip 0.5ex} \hline 
\rowcolor[rgb]{0.95,0.95,0.95} \raisebox{-0.2\totalheight}{\textbf{Final Judgement of LLM Evaluator}} \\ [0.1ex]
\noalign{\vskip 0.5ex} 
Therefore, Output \setlength{\fboxsep}{1.5pt}\colorbox{customred}{B} is better. \\
\bottomrule
\end{tabular}
\caption{Explanations and judgment of pairwise evaluators from an evaluation sample. The \setlength{\fboxsep}{2pt}\colorbox{tab1Blue}{correct} and \setlength{\fboxsep}{2pt}\colorbox{red!30}{incorrect} model explanations are manually highlighted.}
\label{table:headline_evaluation}
\end{table}
\vspace{-0.3cm}

\subsection{Results}
\noindent{\textbf{Pairwise Comparisons Amplifies the Biased Preference.}}\, Fig.~\ref{fig:analysis_comparison} 
shows the meta-evaluation results on the two datasets. 
The pairwise evaluator performs better than the pointwise evaluator on MT-Bench, while the pointwise evaluator outperforms the pairwise evaluator on LLMBar. 
These contrasting results likely stem from the differences in the evaluation structure. 
The pairwise approach excels with normal samples like those in MT-Bench since it can leverage contextual information by considering two outputs simultaneously. However, the benefits of the pairwise approach turn into drawbacks when handling adversarial samples in LLMBar, as it makes the evaluators more easily susceptible to intrinsic biases. In contrast, the pointwise approach is relatively less affected by these biases, resulting in better performance on adversarial samples.

From these observations, we have questioned how pairwise approach undermines the discerning ability of LLM evaluators on adversarial examples. Therefore, our subsequent experiments explore how intrinsic biases impact LLM evaluators using pairwise approach in adversarial evaluations.

\noindent{\textbf{Pairwise Evaluators Do Correct Reasoning behind Incorrect Evaluation.}}\,
To further understand why pairwise evaluation struggles with adversarial samples in LLMBar, we used a Chain-of-Thought~\citep[CoT]{stepbystep_cot} prompting to let LLM provide its own reasoning process before making predictions. 
Specifically, we encouraged the LLM to explain the strengths and weaknesses of each output to see if it can accurately identify issues in low-quality outputs.
Then, we manually examine the explanation provided by LLM for 100 incorrectly predicted samples. More details on human analysis are described in Appendix \ref{sec:app:human_annotation}.

 Our analysis reveals that, despite its incorrect final decisions, the pairwise evaluator often provides rational explanations for identifying flaws in low-quality outputs.
We find that 64 out of 100 explanations accurately describe weaknesses that align with our manual annotations. 
For example, as shown in Table~\ref{table:headline_evaluation}, the pairwise evaluator correctly identifies problems in a flawed output but ultimately makes the wrong judgment. 
This indicates that while the LLM evaluator with the pairwise approach has the potential to perform better, it is hindered by intrinsic issues within the pairwise setup. 
Therefore, addressing these issues is essential for reliable pairwise evaluation.

\section{\textsc{PRePair}: Pointwise Reasoning for Pairwise Evaluation}
\label{sec:method}

\noindent{\textbf{Approach}}
Our analysis in Section~\ref{sec:analysis} revealed two key limitations in existing LLM-based evaluation methods. First, pairwise evaluators often struggle with adversarial samples, as they may still prefer a problematic response despite recognizing its flaws. Second, pointwise evaluators, which analyze responses independently, tend to perform worse on normal samples but show greater robustness against adversarial manipulations. These observations suggest the need for a method that retains the contextual advantages of pairwise evaluation while reducing biases that arise from direct comparison.


\begin{figure}
  \begin{center}
    \vspace{-0.4cm}
    \includegraphics[width=1.0\textwidth]{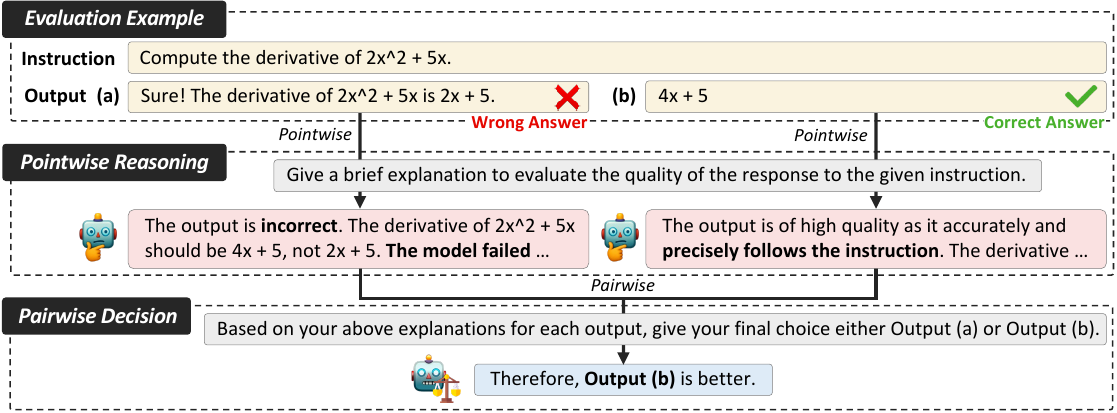}
  \end{center}
  \caption{The overall illustration of \textsc{PRePair}.}
\label{fig:method}
\vspace{-0.3cm}
\end{figure}

To address this, we propose \textsc{PRePair} (Pointwise Reasoning for Pairwise Evaluation), a hybrid approach that incorporates pointwise reasoning into a pairwise evaluation framework. Unlike conventional pairwise evaluation, which directly compares two responses, \textsc{PRePair} first analyzes each response independently using pointwise reasoning before making a final pairwise decision. Given an instruction and two candidate responses, the evaluator reasons about each response separately, considering its properties without direct reference to the other. These independent reasoning steps provide structured insights that are then used in the final pairwise decision, where the evaluator determines the superior response. By decoupling the reasoning process from direct comparison, \textsc{PRePair} helps mitigate biases such as favoring verbosity or model-specific stylistic preferences while maintaining the benefits of pairwise evaluation.

One trade-off of \textsc{PRePair} is the increased computational cost, as it requires an additional LLM forward pass for each response before making a final comparison. However, this overhead can be reduced through caching, since the pointwise reasoning results for individual responses can be reused across multiple comparisons. This is particularly advantageous in ranking tasks, where precomputed reasoning steps allow for efficient and consistent evaluation. By combining the robustness of pointwise reasoning with the contextual awareness of pairwise evaluation, \textsc{PRePair} offers a more reliable evaluation framework, particularly in adversarial scenarios.



\section{Experiments}
\subsection{Setup}
We conduct experiments on the LLMs described in Section~\ref{sec:analysis-setup} and the \texttt{Claude-3-Haiku} \citep{anthropic2024claude}. 
We compare our method for the pairwise approach against the following baselines. 
\textit{w/ CoT} generates an explanation before making a prediction.
\textsc{PRePair}* is an ablated version of our method that prompts pairwise evaluators to generate individual explanations for each output, using the same format as \textsc{PRePair}.
This baseline aims to validate the effectiveness of the pointwise reasoning process.
Detailed setups and more results are in Appendix \ref{sec:app:exp_detail} and \ref{sec:app:additional_results}.

\begin{table*}[ht!]
\small
\centering
    \begin{tabular}{lcccccc}
        \toprule
                                    \noalign{\vskip -0.1ex}
        \textbf{Methods}            & \textbf{GPT-3.5-T}  & \textbf{Claude-3} & \textbf{Llama-3.1} & \textbf{GPT-4} & \textbf{GPT-4o-m} & \textbf{GPT-4o} \\ 
                                    \noalign{\vskip 0.2ex}
        \hline
        \noalign{\vskip 0.5ex}
        Pointwise                   &52.35&	-       & 61.44 &	82.76 & 78.05 & 87.77 \\
        \hline
        \noalign{\vskip 0.5ex}
        Pairwise                    &36.05       &	40.75  & 48.59 &	77.74 & 68.03 & 85.89 \\
        \,\, w/ CoT                 &31.66       &	41.07  & 50.47 &	77.12 & 63.01 & 83.07 \\
        \,\, w/ \textsc{PRePair}\textbf{*}      &\underline{37.62}&	\underline{43.26} & \underline{61.44} &	\underline{78.37} & \underline{68.65} & \textbf{86.21} \\ 
        \,\, w/ \textsc{PRePair}       &\textbf{54.23}&	\textbf{53.60} & \textbf{70.84} &	\textbf{80.88} & \textbf{78.06} & \textbf{86.21} \\
        \noalign{\vskip -0.3ex}
        \bottomrule
\end{tabular}    
        \caption{LLMBar-Adversarial results. The pairwise approach's highest and second-highest scores are highlighted in bold and underlined. \textsc{PRePair}\textbf{*} adopts pairwise reasoning but follows the generation format of \textsc{PRePair}.}
    \label{table:main_llmbar}
\end{table*}

\begin{table*}[ht!]
\centering

\centering
\vspace{-0.2cm}
\begin{tabular}{lcccc}
    \toprule
                                \noalign{\vskip -0.1ex}
    \textbf{Methods}            & \textbf{GPT-3.5-T}  & \textbf{GPT-4o-m} &\textbf{GPT-4}  & \textbf{GPT-4o} \\ 
                                \noalign{\vskip 0.2ex}
    \hline
    \noalign{\vskip 0.5ex}
    Pointwise                                & 51.55 & 61.72 & 54.64&52.29 \\
    \hline
    \noalign{\vskip 0.5ex}
    Pairwise                                 & 60.65 & \textbf{67.20} &\textbf{67.90} &\underline{69.41}  \\
    \,\, w/ CoT                              & \textbf{61.54} & \underline{65.87} &66.40 & 67.46 \\
    \,\, w/ \textsc{PRePair}\textbf{*}       & 60.83 & 64.72 & \underline{66.76}& 69.14\\
    \,\, w/ \textsc{PRePair}                 & \underline{61.36} & 64.10  & 65.34& \textbf{69.50}\\
    \noalign{\vskip -0.3ex}
    \bottomrule
\end{tabular}

\caption{MT-Bench results. The indicators are the same as Table \ref{table:main_llmbar}.}
\label{table:main_mtbench}
\end{table*}
\vspace{-0.3cm}

\subsection{Results}
As shown in Table~\ref{table:main_llmbar}, our method improves the accuracy of pairwise evaluators on the LLMBar-Adversarial dataset across all tested LLMs, with the largest gain observed in Llama-3.1. By incorporating pointwise reasoning, \textsc{PRePair} mitigates biases in direct comparison and enhances robustness against adversarial responses.
A key observation is that \textsc{PRePair} consistently outperforms \textsc{PRePair}\textbf{*}, indicating that its improvements stem from explicit pointwise reasoning rather than structural modifications alone. Additionally, \textsc{PRePair} sometimes performs even better than pointwise evaluators, suggesting that while pointwise reasoning enhances adversarial robustness, integrating it within a pairwise framework preserves the benefits of contextual evaluation. Qualitative examples in Appendix \ref{sec:app:qualitative_prepair} further illustrate how our method improves reliability compared to standard Chain-of-Thought (CoT) prompting.

On the MT-Bench dataset (Table~\ref{table:main_mtbench}), pairwise methods achieve similar performance and consistently outperform the pointwise approach. Since MT-Bench focuses on general response quality rather than adversarial robustness, this result suggests that \textsc{PRePair} does not compromise effectiveness in standard evaluation while offering clear benefits in adversarial settings. Overall, these results demonstrate that \textsc{PRePair} effectively balances the strengths of pointwise and pairwise evaluation, improving reliability against adversarial samples while maintaining strong performance in normal evaluation tasks.

\subsection{Analysis}
We conduct further analysis to understand \textsc{PrePair} comprehensively as follows.

\begin{table*}[!ht]
\centering
\vspace{-0.2cm}
\begin{tabular}{l|cccc|c}
\toprule
\noalign{\vskip -0.2ex}
\textbf{Methods} & \textbf{Manual} & \textbf{GPTInst} & \textbf{GPTOut} & \textbf{Neighbor} & \textbf{Total} \\
\hline
\multicolumn{6}{c}{
\cellcolor[HTML]{F3F3F3} \raisebox{-0.3\totalheight}{\textit{\textbf{Llama-2-13b-chat}}}} \\ [0.5ex] \hline
\noalign{\vskip 0.4ex}
Pairwise & 28.26 & 28.26 & 51.06 & 29.10 & 31.97 \\
\,\,w/ CoT & 32.61 & 35.87 & 57.45 & 22.39 & 32.92 \\
\,\,w/ \textsc{PrePair} & \textbf{36.96} & 32.61 & \textbf{63.83} & \textbf{40.30} & \textbf{41.07} \\ \hline
\multicolumn{6}{c}{
\cellcolor[HTML]{F3F3F3} \raisebox{-0.2\totalheight}{\textit{\textbf{Llama-2-70b-chat}}}} \\ [0.5ex] \hline
\noalign{\vskip 0.4ex}
Pairwise & 34.78 & 28.26 & 48.94 & 22.39 & 29.78 \\
\,\,w/ CoT & 43.48 & 29.35 & 48.94 & 39.55 & 38.56 \\
\,\,w/ \textsc{PrePair} & 39.13 & \textbf{52.17} & 48.94 & 38.81 & \textbf{44.20} \\ \hline
\multicolumn{6}{c}{
\cellcolor[HTML]{F3F3F3} \raisebox{-0.2\totalheight}{\textit{\textbf{Llama-3-8B-Instruct}}}} \\ [0.5ex] \hline
\noalign{\vskip 0.4ex}
Pairwise & 45.65 & 48.91 & 48.94 & 32.09 & 41.38 \\
\,\,w/ CoT & 54.35 & 46.74 & 48.94 & 41.04 & 45.77 \\
\,\,w/ \textsc{PrePair} & 52.17 & \textbf{72.83} & \textbf{51.06} & \textbf{58.96} & \textbf{60.82} \\ \hline
\multicolumn{6}{c}{
\cellcolor[HTML]{F3F3F3} \raisebox{-0.2\totalheight}{\textit{\textbf{Llama-3.1-8B-Instruct}}}} \\ [0.5ex] \hline
\noalign{\vskip 0.4ex}
Pairwise & 60.87 & 73.91 & 61.70 & 52.99 & 61.44 \\
\,\,w/ CoT & 52.17 & 55.43 & 53.19 & 45.52 & 50.47 \\
\,\,w/ \textsc{PrePair} & \textbf{69.57} & \textbf{77.17} & 53.19 & \textbf{73.13} & \textbf{70.84} \\
\bottomrule
\end{tabular}
\caption{Accuracies on the LLMBar-Adversarial dataset for different Llama model variants. Each model is evaluated using Pairwise, Chain-of-Thought (CoT), and PrePair.}
\label{app:tab:llama_comparison}
\vspace{-0.3cm}
\end{table*}

\begin{table*}[!ht]
\centering
\begin{tabular}{l|cccc|c}
\toprule
\noalign{\vskip -0.2ex}
\textbf{Methods} & \textbf{Manual} & \textbf{GPTInst} & \textbf{GPTOut} & \textbf{Neighbor} & \textbf{Total} \\
\hline
\noalign{\vskip 0.4ex}
Pointwise & 47.83 & 54.35 & 55.32 & 43.28 & 48.90 \\
Pairwise & 50.00 & 33.70 & \textbf{57.45} & 32.84 & 39.19 \\
\,\,w/ \textsc{PrePair} & \textbf{58.70} & \textbf{57.61} & \textbf{57.45} & 35.82 & 48.59 \\
\bottomrule
\end{tabular}
\caption{Accuracies on adversarial datasets for Prometheus-v2-7b \citep{kim2024prometheus} model evaluated across different LLMBar-Adversarial subsets.}
\vspace{-0.3cm}
\label{tab:prometheus_v2}
\end{table*}

\subsubsection{Evaluating \textsc{PrePair} to Open-source LLMs}

We use the following open-source LLMs to probe the effectiveness of PrePair: Llama-2-7b/70b-chat \citep{touvron2023llama2}, Llama-3-8B-Instruct \citep{dubey2024llama3}, and Llama-3.1-8b-Instruct \citep{dubey2024llama3}. As shown in Table \ref{app:tab:llama_comparison}, PrePair consistently contributes to better performance of LLM evaluators on adversarial meta-evaluation datasets.

\subsubsection{Evaluating \textsc{PrePair} to a Finetuned LLM Evaluator}
We conducted experiments with Prometheus-v2-7B\footnote{\url{https://huggingface.co/prometheus-eval/prometheus-7b-v2.0}} \citep{kim2024prometheus}, which supports both pointwise and pairwise evaluations.
The evaluation results are shown in Table \ref{tab:prometheus_v2}. We observe consistent trends with our previous results, where a pointwise evaluation usually performs better than a pairwise one in adversarial samples, and PrePair increases the performance of pairwise evaluators.

\begin{table*}[ht!]
\vspace{-0.4cm}
\centering
\small
\begin{tabular}{l|cccc|c}
\toprule
\textbf{Methods}  & \cellcolor[HTML]{FFFFFF}\textbf{Manual} & \cellcolor[HTML]{FFFFFF}\textbf{GPTInst} & \cellcolor[HTML]{FFFFFF}\textbf{GPTOut} & \cellcolor[HTML]{FFFFFF}\textbf{Neighbor} & \cellcolor[HTML]{FFFFFF}\textbf{Total} \\
 \midrule 

Pointwise & \underline{56.52} & 61.96 & 38.30 & 49.25 & 52.35 \\
Pairwise  & 41.30 & 33.70 & 40.43 & 34.33 & 36.05 \\
\,\,w/ CoT  & 28.26 & 39.13 & 46.81 & 22.39 & 31.66 \\
\,\,w/  SGM 	&52.17	&55.43	&46.81	&\underline{55.97}	&53.92 \\
\,\,w/ \textsc{PrePair}  & 52.17 & \textbf{64.13} & \underline{48.94} & 50.00 & \underline{54.23} \\
\,\,w/ \textsc{PrePair} + SGM&	\textbf{58.70}&	\underline{63.04}	&\textbf{51.06}	&\textbf{57.46}&	\textbf{58.31} \\
\bottomrule
\end{tabular}
\caption{LLMBar-Adversarial results with Self-generated Metrics (SGM) \citep{saha2023branch}. }
\label{tab:app:self-generated}
\vspace{-0.2cm}
\end{table*}

\begin{table}[t]
    \vspace{-0.2cm}
    \centering
    \small
    \begin{tabular}{lccccc|c}
        \toprule        
        Model
        & \makecell[c]{AlpacaEval \\-Easy \\ (100)} 
        & \makecell[c]{AlpacaEval \\-Hard \\ (95)}  
        & \makecell[c]{AlpacaEval \\-Length \\ (95)}  
        & \makecell[c]{MTBench \\-Easy \\ (28)}  
        & \makecell[c]{MTBench \\-Med \\ (40)}  
        & \makecell[c]{Total \\ (358)}  \\

        \hline        
        \multicolumn{7}{c}{
\cellcolor[HTML]{F3F3F3} \raisebox{-0.2\totalheight}{\textit{\textbf{GPT-3.5-Turbo}}}} \\ [0.5ex] \hline
        \noalign{\vskip 0.4ex}
        Pointwise & 75.00 & 72.63 & 74.74 & 92.86 & 75.00 & 75.70 \\
        Pairwise & 89.00 & 91.58 & 83.16 & 100.00 & 80.00 & 87.99 \\
        \hline        
        \multicolumn{7}{c}{
\cellcolor[HTML]{F3F3F3} \raisebox{-0.2\totalheight}{\textit{\textbf{GPT-4}}}} \\ [0.5ex] \hline
        \noalign{\vskip 0.4ex}
        Pointwise & 87.00 & 84.21 & 86.32 & 100.00 & 100.00 & 88.55 \\
        Pairwise & 96.00 & 95.79 & 92.63 & 100.00 & 100.00 & 95.81 \\
        \hline
        \multicolumn{7}{c}{
\cellcolor[HTML]{F3F3F3} \raisebox{-0.2\totalheight}{\textit{\textbf{GPT-4o-mini}}}} \\ [0.5ex] \hline
        \noalign{\vskip 0.4ex}
        Pointwise & 93.00 & 88.42 & 89.47 & 100.00 & 97.50 & 91.90 \\
        Pairwise & 96.00 & 94.74 & 96.84 & 100.00 & 95.00 & 96.09 \\ \hline
        \multicolumn{7}{c}{
\cellcolor[HTML]{F3F3F3} \raisebox{-0.2\totalheight}{\textit{\textbf{GPT-4o}}}} \\ [0.5ex] \hline
        \noalign{\vskip 0.4ex}
        Pointwise & 90.00 & 89.47 & 84.21 & 100.00 & 100.00 & 90.22 \\
        Pairwise & 93.00 & 92.63 & 92.63 & 100.00 & 97.50 & 93.85 \\
        \bottomrule
    \end{tabular}
    \caption{Evaluation results of different evaluation approach on the RewardBench-Chat \citep{lambert2024rewardbench}. The pointwise evaluation follows a weighted sum aggregation over score token probabilities, while the pairwise evaluation directly compares two responses.}
    \vspace{-0.4cm}
    \label{tab:rewardbench_results}
\end{table}

\subsubsection{Extending \textsc{PrePair} with Different Reasoning Strategies}
\label{sec:app:self-generated-metric}

To assess the extensibility of \textsc{PRePair}, we apply it to the Self-Generated Metric approach \citep{saha2023branch, zeng2024llmbar}, where LLMs first generate instance-specific evaluation criteria before making judgments. These criteria are then incorporated into prompts to guide the evaluation process.

We compare \textsc{PRePair} with standard pairwise evaluation when generating both metrics and explanations. As shown in Table~\ref{tab:app:self-generated}, \textsc{PRePair} consistently improves performance, suggesting that its reasoning mechanism enhances evaluation robustness. Moreover, its effectiveness across different reasoning setups indicates that \textsc{PRePair} can complement other evaluation strategies, making it a flexible addition to existing frameworks.

\vspace{-0.1cm}
\subsubsection{Impacts of Different Aggregation Strategies in Pointwise Approach}
\label{sec:reward-chat}
In pointwise evaluation, the final score for a response is typically derived through an aggregation strategy, which determines how individual scoring tokens contribute to the overall decision. A common approach is to compute a weighted sum over the probabilities of score tokens, which has been shown to improve performance \citep{liu2023gpteval}. However, certain benchmarks, such as MT-Bench, allow tie predictions, requiring an alternative strategy. To accommodate this, our pointwise evaluator for MT-Bench directly uses the scalar value generated by the model instead of applying a weighted sum. In contrast, the RewardBench-Chat subset \citep{lambert2024rewardbench} does not include tie options, allowing a direct comparison between aggregation strategies. Here, the pointwise evaluator follows the same setup as in LLMBar, computing scores based on a weighted sum over Likert-scale integer tokens. The dataset, curated from AlpacaEval \citep{dubois2024alpacafarm} and MT-Bench, consists of 358 samples, each containing an instruction, two model outputs, and a preference label indicating the superior response without the possibility of a tie.

Table~\ref{tab:rewardbench_results} presents the results, showing that even when using a weighted sum aggregation, pairwise evaluation still outperforms pointwise evaluation. This indicates that the aggregation strategy alone does not bridge the performance gap between the two approaches. These findings reinforce our claim that pointwise evaluation is particularly beneficial in adversarial settings, whereas pairwise evaluation remains more effective in general cases.

\subsubsection{Qualitative Result}
Table~\ref{table:qualitative_main_table} presents a comparison between standard Chain-of-Thought (CoT) reasoning and our \textsc{PRePair}-based evaluation on a sample from the dataset. \texttt{GPT-3.5-Turbo} is used as an LLM evaluator. The evaluation involves assessing two outputs in response to a prompt asking for an illustrative example of artificial intelligence. The standard CoT evaluator favors Output (a) for its breadth and coverage of various AI applications, overlooking the fact that the instruction specifically asks for a single example. In contrast, \textsc{PRePair} provides a more precise analysis, recognizing that Output (a) does not strictly follow the instruction, whereas Output (b) effectively aligns with the task by offering a clear and illustrative example of AI in action. This example highlights how \textsc{PRePair} enhances evaluation by focusing on adherence to instructions rather than surface-level completeness, thereby mitigating biases in pairwise comparison.

\definecolor{customblue}{rgb}{0.812, 0.835, 0.976}
\definecolor{customred}{rgb}{0.961, 0.741, 0.714}

\begin{table}[t!]
\small
\vspace{-0.9cm}
\centering
\begin{adjustbox}{width=\textwidth}
\begin{tabular}{p{\textwidth}}
\bottomrule
\rowcolor[rgb]{0.95,0.95,0.95} \raisebox{-0.2\totalheight}{\textbf{Evaluation Sample}} \\ [0.5ex]
\noalign{\vskip 0.5ex} 
\textbf{Instruction:} Generate an example that illustrates the concept of "artificial intelligence". \\ 
\noalign{\vskip 0.5ex} 
\cdashline{1-1}
\noalign{\vskip 0.5ex}
\textbf{Output A}: Artificial intelligence has numerous applications in real life. Examples include self-driving cars, facial recognition software, and voice assistants such as Siri and Alexa. AI is also used in medical applications such as disease detection, drug development and medical diagnostics. AI is also used in finance applications such as automated trading, fraud detection and predictive analytics.\\ 
\noalign{\vskip 0.5ex}
\textbf{Output B}: An example of artificial intelligence is a robotic vacuum cleaner that is able to map out a room and create an efficient cleaning path. It can also adjust its cleaning strategies based on different levels of dirt and debris. (\textcolor{blue}{\textit{\textbf{Human-preferred}}})\\
\hline 
\rowcolor[rgb]{0.95,0.95,0.95} \raisebox{-0.2\totalheight}{\textbf{Vanilla Chain-of-Thought}} \\ [0.5ex]

\noalign{\vskip 0.5ex} 
\textbf{Explanation of Output (a) and Output (b)} Output (a) provides a broader and more comprehensive illustration of the concept of artificial intelligence by mentioning various real-life applications across different fields such as technology, medicine, and finance. It gives a well- rounded view of the capabilities and impact of AI. \setlength{\fboxsep}{1pt}\colorbox{red!30}{On the other hand, Output (b) focuses on a specific example} of a robotic vacuum cleaner, which, although a good example, lacks the depth and variety of applications highlighted in Output (a). / \textbf{Decision:} \setlength{\fboxsep}{1pt}\colorbox{red!30}{Output (a)} \\

\hline 
\rowcolor[rgb]{0.95,0.95,0.95} \raisebox{-0.2\totalheight}{\textbf{PrePair} (Ours))} \\ [0.5ex]

\noalign{\vskip 0.5ex} 
\textbf{Explanation of Output (a)} The output provided goes beyond the scope of the instruction by listing multiple examples of artificial intelligence applications rather than generating a single example to illustrate the concept of "artificial intelligence." While the examples mentioned are relevant and informative, the response fails to precisely execute the instruction by not focusing on a single illustrative example. To improve, the \setlength{\fboxsep}{1pt}\colorbox{tab1Blue}{AI chatbot should provide a specific and concise example} that clearly demonstrates the concept of artificial intelligence. 

\textbf{Explanation of Output (b)} The output provided does a good job of illustrating the concept of "artificial intelligence" by giving a specific example of a robotic vacuum cleaner that demonstrates AI capabilities. It effectively showcases how AI can be applied in everyday technology to perform tasks autonomously and adaptively. The example is relevant, clear, and directly aligns with the instruction given. Overall, the output is of high quality as \setlength{\fboxsep}{1pt}\colorbox{tab1Blue}{it precisely executes the instruction} by providing a suitable example of artificial intelligence in action. / \textbf{Decision:} \setlength{\fboxsep}{1pt}\colorbox{tab1Blue}{Output (b)} \\

\bottomrule
\end{tabular}
\end{adjustbox}
\caption{Explanations and judgment of different pairwise evaluators from an LLMBar-Adversarial sample. The \setlength{\fboxsep}{1pt}\colorbox{tab1Blue}{correct} and \setlength{\fboxsep}{1pt}\colorbox{red!30}{incorrect} generations are manually highlighted.}
\label{table:qualitative_main_table}
\vspace{-0.4cm}
\end{table}
\section{Discussion}
\label{sec:conclusion}
\vspace{-0.1cm}
In this study, we analyze the reliability of LLM evaluators using two approaches: pairwise (evaluating outputs simultaneously) and pointwise (evaluating outputs individually). Our findings show that in adversarial settings, pairwise evaluators favor superficial attributes with biases. In contrast, while pointwise evaluators improve performance in those settings, they struggle in normal settings.
Based on these findings, we propose a hybrid approach, \textsc{PRePair}, which analyzes outputs pointwise before making a final pairwise decision.
We hope our work offers insights for improving LLM evaluators.
Despite its efficacy, however, \textsc{PRePair} may not be the ultimate solution for LLM-based evaluations. 
The ultimate goal should be to enable LLMs to discern response quality based on human preference hierarchies, even when presented with adversarial samples simultaneously. 
Nevertheless, reaching this goal is challenging, as even powerful proprietary models have yet to achieve this. 
Therefore, our work is important for advancing the development of LLM evaluators that better align with human judgment.


\bibliography{colm2025_conference}

\begin{thebibliography}{35}
\providecommand{\natexlab}[1]{#1}
\providecommand{\url}[1]{\texttt{#1}}
\expandafter\ifx\csname urlstyle\endcsname\relax
  \providecommand{\doi}[1]{doi: #1}\else
  \providecommand{\doi}{doi: \begingroup \urlstyle{rm}\Url}\fi

\bibitem[Achiam et~al.(2023)Achiam, Adler, Agarwal, Ahmad, Akkaya, Aleman, Almeida, Altenschmidt, Altman, Anadkat, et~al.]{gpt4}
Josh Achiam, Steven Adler, Sandhini Agarwal, Lama Ahmad, Ilge Akkaya, Florencia~Leoni Aleman, Diogo Almeida, Janko Altenschmidt, Sam Altman, Shyamal Anadkat, et~al.
\newblock Gpt-4 technical report.
\newblock \emph{arXiv preprint arXiv:2303.08774}, 2023.

\bibitem[Anthropic(2024)]{anthropic2024claude}
AI~Anthropic.
\newblock The claude 3 model family: Opus, sonnet, haiku.
\newblock \emph{Claude-3 Model Card}, 2024.

\bibitem[Banerjee \& Lavie(2005)Banerjee and Lavie]{banerjee-lavie-2005-meteor}
Satanjeev Banerjee and Alon Lavie.
\newblock {METEOR}: An automatic metric for {MT} evaluation with improved correlation with human judgments.
\newblock In \emph{Proceedings of the {ACL} Workshop on Intrinsic and Extrinsic Evaluation Measures for Machine Translation and/or Summarization}, pp.\  65--72, Ann Arbor, Michigan, June 2005. Association for Computational Linguistics.
\newblock URL \url{https://aclanthology.org/W05-0909}.

\bibitem[Chan et~al.(2024)Chan, Chen, Su, Yu, Xue, Zhang, Fu, and Liu]{chan2023chateval}
Chi-Min Chan, Weize Chen, Yusheng Su, Jianxuan Yu, Wei Xue, Shanghang Zhang, Jie Fu, and Zhiyuan Liu.
\newblock Chateval: Towards better llm-based evaluators through multi-agent debate.
\newblock In \emph{The Twelfth International Conference on Learning Representations}, 2024.

\bibitem[Chen et~al.(2024)Chen, Chen, Liu, Jiang, and Wang]{chen2024humans}
Guiming~Hardy Chen, Shunian Chen, Ziche Liu, Feng Jiang, and Benyou Wang.
\newblock Humans or llms as the judge? a study on judgement biases.
\newblock \emph{arXiv preprint arXiv:2402.10669}, 2024.

\bibitem[Chiang \& Lee(2023)Chiang and Lee]{chiang2023can}
Cheng-Han Chiang and Hung-yi Lee.
\newblock Can large language models be an alternative to human evaluations?
\newblock In Anna Rogers, Jordan Boyd-Graber, and Naoaki Okazaki (eds.), \emph{Proceedings of the 61st Annual Meeting of the Association for Computational Linguistics (Volume 1: Long Papers)}, pp.\  15607--15631, Toronto, Canada, July 2023. Association for Computational Linguistics.
\newblock \doi{10.18653/v1/2023.acl-long.870}.
\newblock URL \url{https://aclanthology.org/2023.acl-long.870}.

\bibitem[Chiang et~al.(2024)Chiang, Zheng, Sheng, Angelopoulos, Li, Li, Zhu, Zhang, Jordan, Gonzalez, et~al.]{chiang2024chatbot}
Wei-Lin Chiang, Lianmin Zheng, Ying Sheng, Anastasios~Nikolas Angelopoulos, Tianle Li, Dacheng Li, Banghua Zhu, Hao Zhang, Michael Jordan, Joseph~E Gonzalez, et~al.
\newblock Chatbot arena: An open platform for evaluating llms by human preference.
\newblock In \emph{Forty-first International Conference on Machine Learning}, 2024.

\bibitem[Dubey et~al.(2024)Dubey, Jauhri, Pandey, Kadian, Al-Dahle, Letman, Mathur, Schelten, Yang, Fan, et~al.]{dubey2024llama3}
Abhimanyu Dubey, Abhinav Jauhri, Abhinav Pandey, Abhishek Kadian, Ahmad Al-Dahle, Aiesha Letman, Akhil Mathur, Alan Schelten, Amy Yang, Angela Fan, et~al.
\newblock The llama 3 herd of models.
\newblock \emph{arXiv preprint arXiv:2407.21783}, 2024.

\bibitem[Dubois et~al.(2024)Dubois, Li, Taori, Zhang, Gulrajani, Ba, Guestrin, Liang, and Hashimoto]{dubois2024alpacafarm}
Yann Dubois, Chen~Xuechen Li, Rohan Taori, Tianyi Zhang, Ishaan Gulrajani, Jimmy Ba, Carlos Guestrin, Percy~S Liang, and Tatsunori~B Hashimoto.
\newblock Alpacafarm: A simulation framework for methods that learn from human feedback.
\newblock \emph{Advances in Neural Information Processing Systems}, 36, 2024.

\bibitem[Einhorn \& Hogarth(1981)Einhorn and Hogarth]{einhorn1981behavioral}
Hillel~J Einhorn and Robin~M Hogarth.
\newblock Behavioral decision theory: Processes of judgment and choice.
\newblock \emph{Journal of Accounting Research}, pp.\  1--31, 1981.

\bibitem[Kim et~al.(2024)Kim, Suk, Longpre, Lin, Shin, Welleck, Neubig, Lee, Lee, and Seo]{kim2024prometheus}
Seungone Kim, Juyoung Suk, Shayne Longpre, Bill~Yuchen Lin, Jamin Shin, Sean Welleck, Graham Neubig, Moontae Lee, Kyungjae Lee, and Minjoon Seo.
\newblock Prometheus 2: An open source language model specialized in evaluating other language models.
\newblock \emph{arXiv preprint arXiv:2405.01535}, 2024.

\bibitem[Kocmi \& Federmann(2023)Kocmi and Federmann]{kocmi2023nmt-qe-llm}
Tom Kocmi and Christian Federmann.
\newblock Large language models are state-of-the-art evaluators of translation quality.
\newblock In \emph{Proceedings of the 24th Annual Conference of the European Association for Machine Translation}, pp.\  193--203, 2023.

\bibitem[Kojima et~al.(2022)Kojima, Gu, Reid, Matsuo, and Iwasawa]{stepbystep_cot}
Takeshi Kojima, Shixiang~Shane Gu, Machel Reid, Yutaka Matsuo, and Yusuke Iwasawa.
\newblock Large language models are zero-shot reasoners.
\newblock In \emph{ICML 2022 Workshop on Knowledge Retrieval and Language Models}, 2022.

\bibitem[Koo et~al.(2024)Koo, Lee, Raheja, Park, Kim, and Kang]{koo2023benchmarking}
Ryan Koo, Minhwa Lee, Vipul Raheja, Jong~Inn Park, Zae~Myung Kim, and Dongyeop Kang.
\newblock Benchmarking cognitive biases in large language models as evaluators.
\newblock In Lun-Wei Ku, Andre Martins, and Vivek Srikumar (eds.), \emph{Findings of the Association for Computational Linguistics ACL 2024}, pp.\  517--545, Bangkok, Thailand and virtual meeting, August 2024. Association for Computational Linguistics.
\newblock \doi{10.18653/v1/2024.findings-acl.29}.
\newblock URL \url{https://aclanthology.org/2024.findings-acl.29}.

\bibitem[Lambert et~al.(2024)Lambert, Pyatkin, Morrison, Miranda, Lin, Chandu, Dziri, Kumar, Zick, Choi, et~al.]{lambert2024rewardbench}
Nathan Lambert, Valentina Pyatkin, Jacob Morrison, LJ~Miranda, Bill~Yuchen Lin, Khyathi Chandu, Nouha Dziri, Sachin Kumar, Tom Zick, Yejin Choi, et~al.
\newblock Rewardbench: Evaluating reward models for language modeling.
\newblock \emph{arXiv preprint arXiv:2403.13787}, 2024.

\bibitem[Li et~al.(2023)Li, Patel, and Du]{li2023prd}
Ruosen Li, Teerth Patel, and Xinya Du.
\newblock Prd: Peer rank and discussion improve large language model based evaluations, 2023.

\bibitem[Lin(2004)]{lin-2004-rouge}
Chin-Yew Lin.
\newblock {ROUGE}: A package for automatic evaluation of summaries.
\newblock In \emph{Text Summarization Branches Out}, pp.\  74--81, Barcelona, Spain, July 2004. Association for Computational Linguistics.
\newblock URL \url{https://aclanthology.org/W04-1013}.

\bibitem[Lin \& Chen(2023)Lin and Chen]{lin2023llm}
Yen-Ting Lin and Yun-Nung Chen.
\newblock Llm-eval: Unified multi-dimensional automatic evaluation for open-domain conversations with large language models.
\newblock In \emph{Proceedings of the 5th Workshop on NLP for Conversational AI (NLP4ConvAI 2023)}, pp.\  47--58, 2023.

\bibitem[Liu et~al.(2016)Liu, Lowe, Serban, Noseworthy, Charlin, and Pineau]{liu-etal-2016-evaluate}
Chia-Wei Liu, Ryan Lowe, Iulian Serban, Mike Noseworthy, Laurent Charlin, and Joelle Pineau.
\newblock How {NOT} to evaluate your dialogue system: An empirical study of unsupervised evaluation metrics for dialogue response generation.
\newblock In Jian Su, Kevin Duh, and Xavier Carreras (eds.), \emph{Proceedings of the 2016 Conference on Empirical Methods in Natural Language Processing}, pp.\  2122--2132, Austin, Texas, November 2016. Association for Computational Linguistics.
\newblock \doi{10.18653/v1/D16-1230}.
\newblock URL \url{https://aclanthology.org/D16-1230}.

\bibitem[Liu et~al.(2023)Liu, Iter, Xu, Wang, Xu, and Zhu]{liu2023gpteval}
Yang Liu, Dan Iter, Yichong Xu, Shuohang Wang, Ruochen Xu, and Chenguang Zhu.
\newblock {G}-eval: {NLG} evaluation using gpt-4 with better human alignment.
\newblock In Houda Bouamor, Juan Pino, and Kalika Bali (eds.), \emph{Proceedings of the 2023 Conference on Empirical Methods in Natural Language Processing}, pp.\  2511--2522, Singapore, December 2023. Association for Computational Linguistics.
\newblock \doi{10.18653/v1/2023.emnlp-main.153}.
\newblock URL \url{https://aclanthology.org/2023.emnlp-main.153}.

\bibitem[Liusie et~al.(2024)Liusie, Manakul, and Gales]{liusie2024llm}
Adian Liusie, Potsawee Manakul, and Mark Gales.
\newblock Llm comparative assessment: Zero-shot nlg evaluation through pairwise comparisons using large language models.
\newblock In \emph{Proceedings of the 18th Conference of the European Chapter of the Association for Computational Linguistics (Volume 1: Long Papers)}, pp.\  139--151, 2024.

\bibitem[Min et~al.(2023)Min, Krishna, Lyu, Lewis, Yih, Koh, Iyyer, Zettlemoyer, and Hajishirzi]{min-etal-2023-factscore}
Sewon Min, Kalpesh Krishna, Xinxi Lyu, Mike Lewis, Wen-tau Yih, Pang Koh, Mohit Iyyer, Luke Zettlemoyer, and Hannaneh Hajishirzi.
\newblock {FA}ct{S}core: Fine-grained atomic evaluation of factual precision in long form text generation.
\newblock In Houda Bouamor, Juan Pino, and Kalika Bali (eds.), \emph{Proceedings of the 2023 Conference on Empirical Methods in Natural Language Processing}, pp.\  12076--12100, Singapore, December 2023. Association for Computational Linguistics.
\newblock \doi{10.18653/v1/2023.emnlp-main.741}.
\newblock URL \url{https://aclanthology.org/2023.emnlp-main.741}.

\bibitem[Olshavsky(1979)]{olshavsky1979task}
Richard~W Olshavsky.
\newblock Task complexity and contingent processing in decision making: A replication and extension.
\newblock \emph{Organizational behavior and human performance}, 24\penalty0 (3):\penalty0 300--316, 1979.

\bibitem[Papineni et~al.(2002)Papineni, Roukos, Ward, and Zhu]{papineni-etal-2002-bleu}
Kishore Papineni, Salim Roukos, Todd Ward, and Wei-Jing Zhu.
\newblock {B}leu: a method for automatic evaluation of machine translation.
\newblock In \emph{Proceedings of the 40th Annual Meeting of the Association for Computational Linguistics}, pp.\  311--318, Philadelphia, Pennsylvania, USA, July 2002. Association for Computational Linguistics.
\newblock \doi{10.3115/1073083.1073135}.
\newblock URL \url{https://aclanthology.org/P02-1040}.

\bibitem[Payne(1976)]{payne1976task}
John~W Payne.
\newblock Task complexity and contingent processing in decision making: An information search and protocol analysis.
\newblock \emph{Organizational behavior and human performance}, 16\penalty0 (2):\penalty0 366--387, 1976.

\bibitem[Saha et~al.(2024)Saha, Levy, Celikyilmaz, Bansal, Weston, and Li]{saha2023branch}
Swarnadeep Saha, Omer Levy, Asli Celikyilmaz, Mohit Bansal, Jason Weston, and Xian Li.
\newblock Branch-solve-merge improves large language model evaluation and generation.
\newblock In Kevin Duh, Helena Gomez, and Steven Bethard (eds.), \emph{Proceedings of the 2024 Conference of the North American Chapter of the Association for Computational Linguistics: Human Language Technologies (Volume 1: Long Papers)}, pp.\  8352--8370, Mexico City, Mexico, June 2024. Association for Computational Linguistics.
\newblock \doi{10.18653/v1/2024.naacl-long.462}.
\newblock URL \url{https://aclanthology.org/2024.naacl-long.462}.

\bibitem[Sai et~al.(2022)Sai, Mohankumar, and Khapra]{sai2022survey}
Ananya~B Sai, Akash~Kumar Mohankumar, and Mitesh~M Khapra.
\newblock A survey of evaluation metrics used for nlg systems.
\newblock \emph{ACM Computing Surveys (CSUR)}, 55\penalty0 (2):\penalty0 1--39, 2022.

\bibitem[Touvron et~al.(2023)Touvron, Martin, Stone, Albert, Almahairi, Babaei, Bashlykov, Batra, Bhargava, Bhosale, et~al.]{touvron2023llama2}
Hugo Touvron, Louis Martin, Kevin Stone, Peter Albert, Amjad Almahairi, Yasmine Babaei, Nikolay Bashlykov, Soumya Batra, Prajjwal Bhargava, Shruti Bhosale, et~al.
\newblock Llama 2: Open foundation and fine-tuned chat models.
\newblock \emph{arXiv preprint arXiv:2307.09288}, 2023.

\bibitem[Wang et~al.(2024{\natexlab{a}})Wang, Li, Chen, Cai, Zhu, Lin, Cao, Kong, Liu, Liu, and Sui]{Wang2023LargeLM}
Peiyi Wang, Lei Li, Liang Chen, Zefan Cai, Dawei Zhu, Binghuai Lin, Yunbo Cao, Lingpeng Kong, Qi~Liu, Tianyu Liu, and Zhifang Sui.
\newblock Large language models are not fair evaluators.
\newblock In Lun-Wei Ku, Andre Martins, and Vivek Srikumar (eds.), \emph{Proceedings of the 62nd Annual Meeting of the Association for Computational Linguistics (Volume 1: Long Papers)}, pp.\  9440--9450, Bangkok, Thailand, August 2024{\natexlab{a}}. Association for Computational Linguistics.
\newblock \doi{10.18653/v1/2024.acl-long.511}.
\newblock URL \url{https://aclanthology.org/2024.acl-long.511}.

\bibitem[Wang et~al.(2024{\natexlab{b}})Wang, Kulikov, Golovneva, Yu, Yuan, Dwivedi-Yu, Pang, Fazel-Zarandi, Weston, and Li]{wang2024self}
Tianlu Wang, Ilia Kulikov, Olga Golovneva, Ping Yu, Weizhe Yuan, Jane Dwivedi-Yu, Richard~Yuanzhe Pang, Maryam Fazel-Zarandi, Jason Weston, and Xian Li.
\newblock Self-taught evaluators.
\newblock \emph{arXiv preprint arXiv:2408.02666}, 2024{\natexlab{b}}.

\bibitem[Wright(1974)]{wright1974harassed}
Peter Wright.
\newblock The harassed decision maker: Time pressures, distractions, and the use of evidence.
\newblock \emph{Journal of applied psychology}, 59\penalty0 (5):\penalty0 555, 1974.

\bibitem[Zeng et~al.(2024)Zeng, Yu, Gao, Meng, Goyal, and Chen]{zeng2024llmbar}
Zhiyuan Zeng, Jiatong Yu, Tianyu Gao, Yu~Meng, Tanya Goyal, and Danqi Chen.
\newblock Evaluating large language models at evaluating instruction following.
\newblock In \emph{International Conference on Learning Representations (ICLR)}, 2024.

\bibitem[Zhang et~al.(2023)Zhang, Yu, Yu, Lv, Liu, Huang, Xu, and Li]{zhang2023wider}
Xinghua Zhang, Bowen Yu, Haiyang Yu, Yangyu Lv, Tingwen Liu, Fei Huang, Hongbo Xu, and Yongbin Li.
\newblock Wider and deeper llm networks are fairer llm evaluators.
\newblock \emph{arXiv preprint arXiv:2308.01862}, 2023.

\bibitem[Zheng et~al.(2024)Zheng, Chiang, Sheng, Zhuang, Wu, Zhuang, Lin, Li, Li, Xing, et~al.]{zheng2024judging}
Lianmin Zheng, Wei-Lin Chiang, Ying Sheng, Siyuan Zhuang, Zhanghao Wu, Yonghao Zhuang, Zi~Lin, Zhuohan Li, Dacheng Li, Eric Xing, et~al.
\newblock Judging llm-as-a-judge with mt-bench and chatbot arena.
\newblock \emph{Advances in Neural Information Processing Systems}, 36, 2024.

\bibitem[Zhu et~al.(2024)Zhu, Zhao, Chen, Wang, and Xie]{zhu2024promptbench}
Kaijie Zhu, Qinlin Zhao, Hao Chen, Jindong Wang, and Xing Xie.
\newblock Promptbench: A unified library for evaluation of large language models.
\newblock \emph{Journal of Machine Learning Research}, 25\penalty0 (254):\penalty0 1--22, 2024.

\end{thebibliography}
\bibliographystyle{colm2025_conference}

\newpage
\appendix
\definecolor{customyellow}{rgb}{1, 1, 0.3294}

\section{Experimental Details}
\label{sec:app:exp_detail}



\subsection{Models}
We use \texttt{gpt-3.5-turbo-0125}, \texttt{claude-3-} \texttt{haiku-20240307}, \texttt{gpt-4-0613}, \texttt{gpt-4o-mini-} \texttt{2024-07-18}, and \texttt{gpt-4o}\texttt{-2024-08-06}  for GPT-3.5-Turbo, Claude-3-Haiku, GPT-4, GPT-4o-mini, and GPT-4o, respectively. 
The generation temperature is set to 0 for all models in Table \ref{table:main_llmbar} and Table \ref{table:main_mtbench}.
The evaluation results with a different temperature (i.e., 0.7) are also presented in Table \ref{app:tab:full_llmbar}.

\subsection{Implementation Details in LLM Evaluators}
\label{sec:app:eval_impl_detail}
We present further implementation details of different evaluation approaches used in our experiments.

\paragraph{Pointwise} For the evaluation in MT-Bench, we follow the original prompt in \citet{zheng2024judging} that asks LLMs to choose one of the three options (i.e., \textit{"Output A is better"}, \textit{"Output B is better"}, or \textit{"tie"}). For LLMBar, we ask LLMs to evaluate the output in a 5-point Likert scale and obtain a weighted sum between each integer token and its predicted probability \citep{liu2023gpteval}.
Since Claude API does not support the predictive probability of each token, we do not apply this weighted-sum strategy of pointwise evaluator for Claude.
Note that \textsc{PRePair} does not require the token probability as it only performs pointwise reasoning instead of pointwise final judgments.

\paragraph{Pairwise} For the evaluation in MT-Bench, we use the original prompt released by the authors. For the evaluation in LLMBar, we use the \textit{Rules} prompt that specifies general rules for LLM evaluation (e.g., \textit{"prioritize the honest instruction-following"}) as recommended by \citet{zeng2024llmbar}. These general rules are specified in the prompts of both pointwise and pairwise evaluators.

\subsection{Prompt Details}
We present the prompts of different evaluation strategies in Fig. \ref{fig:pros_and_cons} to Fig. \ref{fig:mtbench_prepair_hat}. 

\subsection{Human annotation Details}
\label{sec:app:human_annotation}
The goal of human annotation is to quantitatively assess the LLM's understanding of the shortcomings in the answers it incorrectly identified as correct.\footnote{The CoT prompting used in this analysis improved the pairwise evaluation performance from 36.05 to 40.12 but remained significantly lower than the pointwise evaluation (52.35).} 
Authors manually examined the explanation provided by the LLM for 100 randomly sampled adversarial examples with incorrect predictions.\footnote{Two of the authors with a bachelor's degree or higher participated in this annotation process.}
The examination comprehensively considered the strengths and weaknesses of both the correct and incorrect outputs. However, its reliability may be affected by focusing on the cons of low-quality outputs to assess alignment. 
A case study of the annotation is in Appendix \ref{sec:app:human_annot}.

\section{Additional Results}
\label{sec:app:additional_results}

\subsection{Results with Different Generation Setup}
\label{sec:app:temperature_0.7}

We conduct experiments with reasoning-based evaluation methods when the decoding temperature is  0.7, as reported in  \citet{stepbystep_cot}.
Results in Table \ref{app:tab:full_llmbar} demonstrate the effectiveness of \textsc{PRePair} in different generation configurations.

\subsection{Results with Paraphrased Text Prompt}
Table \ref{tab:gpt35_turbo_paraphrased} presents the evaluation results using paraphrased prompts generated by GPT-4. Despite using these variations, PrePair maintains its effectiveness even in adversarial datasets.

\section{Qualitative Results}
\subsection{Qualitative Results of \textsc{PRePair}}
\label{sec:app:qualitative_prepair}
We present several case studies utilizing our method in Fig.~\ref{fig:gold_2} to \ref{fig:gold_3}. The examples are generated by \texttt{gpt-3.5-turbo-0125} and \texttt{gpt-4o}.

\subsection{Case Study of Human Annotation}
\label{sec:app:human_annot}
In this section, we examine LLM evaluators' capability to identify the shortcomings in the answers it has incorrectly selected as good. 
The manual for examination covers three main considerations:
\begin{enumerate}
    \item \textbf{Existence of Superficial Preference}: We have examined whether the LLM evaluator has demonstrated a preference for responses that, despite failing to follow instructions, presented a superficial form that the LLM tended to favor.
    \item \textbf{Validity of Strengths and Weaknesses}: We have critically assessed whether the evaluator's judgments of each output’s strengths and weaknesses are reasonable and reflect the true quality of the responses.
    \item \textbf{Failure to Follow Instructions}: Most importantly, we have focused on whether the LLM evaluator has consistently recognized "failing to follow instructions" as the key flaw, as this was the primary focus of the examination.
\end{enumerate}
This manual has served as a guide throughout the entire examination process.

As a result, Fig.~\ref{fig:annot_2} to Fig.~\ref{fig:annot_10} show that LLM evaluators effectively identify the critical drawbacks of instruction following failures in responses they rated highly but are of low quality. This drawbacks are highlighted in \colorbox{customyellow}{yellow}.

\begin{table*}[!ht]
\centering
\small
\begin{adjustbox}{width=\textwidth}
\begin{tabular}{lc|cccc|c}
\toprule
 &  & \multicolumn{5}{c}{\textbf{Adversarial (319)}} \\ 
 \multirow{-2}{*}{\textbf{Methods}} & \multirow{-2}{*}{\textbf{Natural (100)}} & \cellcolor[HTML]{FFFFFF}\textbf{Manual (46)} & \cellcolor[HTML]{FFFFFF}\textbf{GPTInst  (92)} & \cellcolor[HTML]{FFFFFF}\textbf{GPTOut (47)} & \cellcolor[HTML]{FFFFFF}\textbf{Neighbor (134)} & \cellcolor[HTML]{FFFFFF}\textbf{Total (319)} \\
 \hline

\multicolumn{7}{c}{
\cellcolor[HTML]{F3F3F3} \raisebox{-0.2\totalheight}{\textit{\textbf{GPT-3.5-Turbo w/ T=0.0 (Greedy Decoding)}}}} \\ [0.5ex]
Pointwise & 81.00 & 56.52 & 61.96 & 38.30 & 49.25 & 52.35 \\
Pairwise & 82.00 & 41.30 & 33.70 & 40.43 & 34.33 & 36.05 \\
\,\,w/ CoT & 83.00 & 28.26 & 39.13 & 46.81 & 22.39 & 31.66 \\
\,\,w/ \textsc{PrePair}\textbf{*} & 82.00 & 36.96 & 41.30 & 46.81 & 32.09 & 37.62 \\
\,\,w/ \textsc{PrePair} & 81.00 & 52.17 & 64.13 & 48.94 & 50.00 & 54.23 \\
\multicolumn{7}{c}{
\cellcolor[HTML]{F3F3F3}\raisebox{-0.2\totalheight}{\textit{\textbf{GPT-3.5-Turbo w/ T=0.7}}}} \\ [0.5ex]
\,\,w/ CoT & 84.00 & 30.43 & 31.52 & 44.68 & 24.63 & 30.41 \\
\,\,w/ \textsc{PrePair}\textbf{*} & 78.00 & 34.78 & 44.57 & 46.81 & 28.36 & 36.68 \\
\,\,w/ \textsc{PrePair} & 81.00 & 54.35 & 61.96 & 57.45 & 54.48 & 57.06 \\
\multicolumn{7}{c}{
\cellcolor[HTML]{F3F3F3}\raisebox{-0.2\totalheight}{\textit{\textbf{GPT-4 w/ T=0.0 (Greedy Decoding)}}}} \\ [0.5ex]
Pointwise & 93.00 & 84.78 & 90.22 & 74.47 & 79.85 & 82.76 \\
Pairwise & 95.00 & 71.74 & 84.78 & 74.47 & 76.12 & 77.74 \\
\,\,w/ CoT & 94.00 & 76.09 & 85.87 & 74.47 & 72.39 & 77.12 \\
\,\,w/ \textsc{PrePair}\textbf{*} & 94.00 & 80.43 & 84.78 & 74.47 & 74.63 & 78.37 \\
\,\,w/ \textsc{PrePair} & 93.00 & 80.43 & 83.70 & 78.72 & 79.85 & 80.88 \\
\multicolumn{7}{c}{
\cellcolor[HTML]{F3F3F3}\raisebox{-0.2\totalheight}{\textit{\textbf{GPT-4 w/ T=0.7}}}} \\ [0.5ex]
\,\,w/ CoT & 95.00 & 76.09 & 86.96 & 76.60 & 70.90 & 77.12 \\
\,\,w/ \textsc{PrePair}\textbf{*} & 95.00 & 73.91 & 86.96 & 74.47 & 76.87 & 79.00 \\
\,\,w/ \textsc{PrePair} & 92.00 & 71.74 & 83.70 & 78.72 & 82.84 & 80.88 \\
\multicolumn{7}{c}{
\cellcolor[HTML]{F3F3F3}\raisebox{-0.2\totalheight}{\textit{\textbf{Claude-3-Haiku w/ T=0.0 (Greedy Decoding)}}}} \\ [0.5ex]
Pairwise & 87.00 & 52.17 & 46.74 & 55.32 & 27.61 & 40.75 \\
\,\,w/ CoT & 87.00 & 43.48 & 45.65 & 59.57 & 30.60 & 41.07 \\
\,\,w/ \textsc{PrePair}\textbf{*} & 81.00 & 43.48 & 45.65 & 65.96 & 33.58 & 43.26 \\
\,\,w/ \textsc{PrePair} & 85.00 & 60.87 & 52.17 & 59.57 & 50.00 & 53.60 \\

\multicolumn{7}{c}{
\cellcolor[HTML]{F3F3F3}\raisebox{-0.2\totalheight}{\textit{\textbf{GPT-4o-mini w/ T=0.0 (Greedy Decoding)}}}} \\ [0.5ex]
Pointwise& 91.00 & 76.09 & 88.04 & 61.70 & 77.61 & 78.05\\
Pairwise & 87.00 & 60.87 & 80.43 & 65.96 & 62.69 & 68.03\\
\,\,w/ CoT & 88.00 & 58.70 & 72.83 & 65.96 & 56.72 & 63.01\\
\,\,w/ PrePair\textbf{*} & 88.00 & 67.39 & 82.61 & 74.47 & 57.46 & 68.65\\
\,\,w/ PrePair & 89.00 & 69.57 & 91.30 & 70.21 & 74.63 & 78.06\\
\multicolumn{7}{c}{
\cellcolor[HTML]{F3F3F3}\raisebox{-0.2\totalheight}{\textit{\textbf{GPT-4o w/ T=0.0 (Greedy Decoding)}}}} \\ [0.5ex]
Pointwise& 96.00&  91.30&  92.39&  78.72&  86.57&  87.77\\
Pairwise & 97.00 & 80.43 & 93.48 & 80.85 & 84.33 & 85.89\\
\,\,w/ CoT & 96.00 & 82.61 & 90.22 & 70.21 & 82.84 & 83.07\\
\,\,w/ PrePair\textbf{*} & 98.00 & 84.78 & 92.39 & 82.98 & 83.58 & 86.21\\
\,\,w/ PrePair & 95.00 & 89.13 & 91.30 & 78.72 & 84.33 & 86.21\\
\multicolumn{7}{c}{
\cellcolor[HTML]{F3F3F3}\raisebox{-0.2\totalheight}{\textit{\textbf{Llama-3.1-8B-Instruct						 w/ T=0.0 (Greedy Decoding)}}}} \\ [0.5ex]
Pointwise& 84.00&  60.87&  73.91&  61.70&  52.99&  61.44\\
Pairwise & 83.00 & 54.35 & 58.70 & 46.81 & 40.30 & 48.59\\
\,\,w/ CoT & 75.00 & 52.17 & 55.43 & 53.19 & 45.52 & 50.47\\
\,\,w/ PrePair\textbf{*} & 80.00 & 63.04 & 61.96 & 51.06 & 64.18 & 61.44\\
\,\,w/ PrePair & 79.00 & 69.57 & 77.17 & 53.19 & 73.13 & 70.84\\

\bottomrule
\end{tabular}
\end{adjustbox}
\caption{LLMBar results with a breakdown of performance in each subset. Numbers in the parenthesis denote the number of evaluation samples in each subset. The values in the Total column represent the weighted average, accounting for the different numbers of examples in each adversarial data type.}
\label{app:tab:full_llmbar}
\end{table*}
\begin{table*}[!ht]
\centering
\small
\begin{tabular}{l|cccc|c}
\toprule
\noalign{\vskip -0.5ex}
\textbf{Methods} & \textbf{Manual (46)} & \textbf{GPTInst (92)} & \textbf{GPTOut (47)} & \textbf{Neighbor (134)} & \textbf{Total (319)} \\
\hline
Pointwise & 45.65 & 48.91 & 42.55 & 45.52 & 46.08 \\
Pairwise & 39.13 & 44.57 & 42.55 & 41.79 & 42.32 \\
\,\,w/ CoT & 32.61 & 43.48 & 44.68 & 13.43 & 29.47 \\
\,\,w/ PrePair \textbf{*}  & 28.26 & 44.57 & 40.43 & 30.60 & 35.74 \\
\,\,w/ PrePair & 47.83 & 71.74 & 51.06 & 58.96 & 59.88 \\
\bottomrule
\end{tabular}
\caption{Accuracies on adversarial datasets for GPT-3.5-Turbo with paraphrased text prompts.}
\label{tab:gpt35_turbo_paraphrased}
\end{table*}

\begin{figure*}[ht]
\centering
\includegraphics[width=\textwidth]{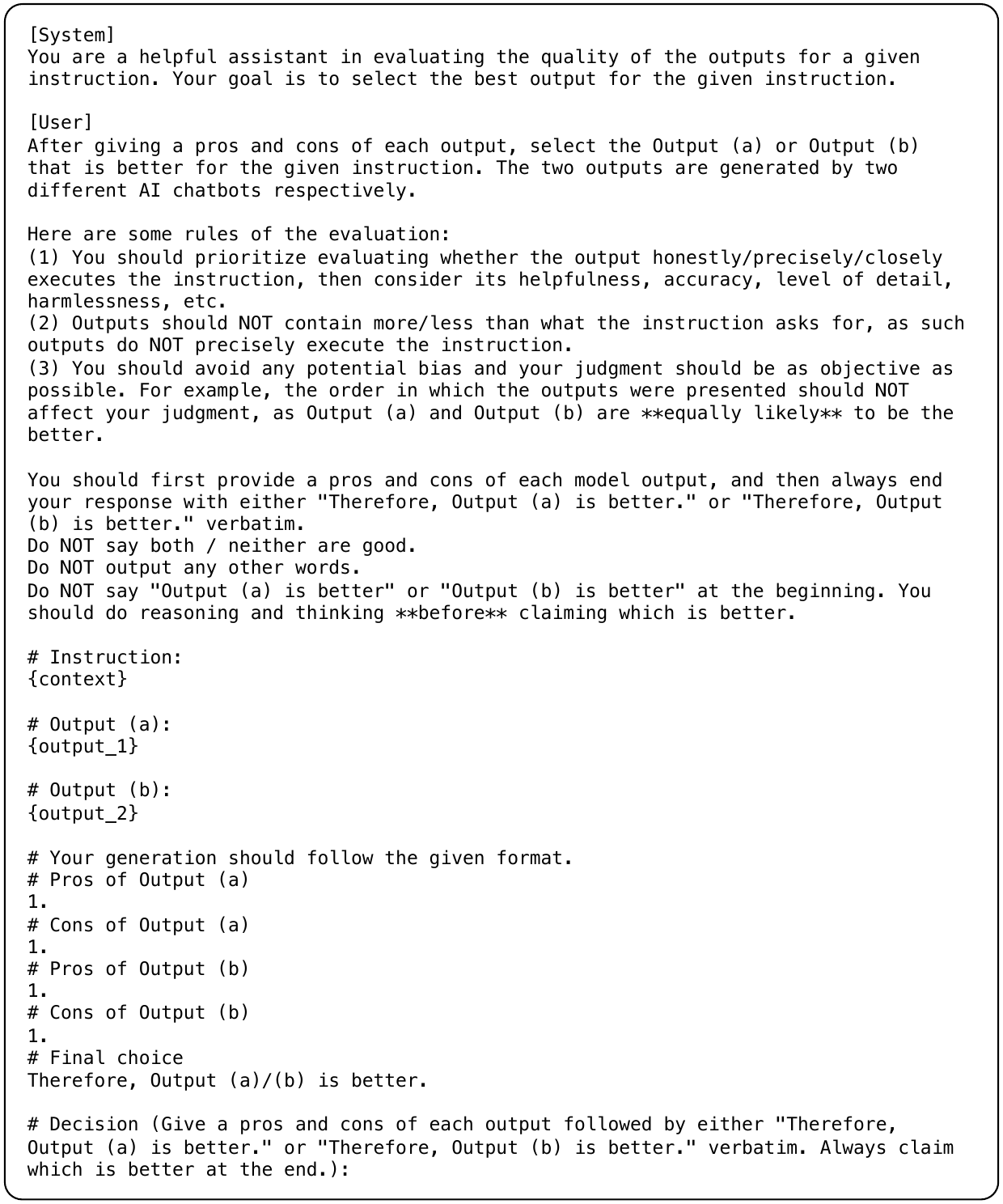}

\caption{The strengths and weaknesses prompt for pairwise evaluation.}
\label{fig:pros_and_cons}
\end{figure*}

\begin{figure*}
\centering
\includegraphics[width=\textwidth]{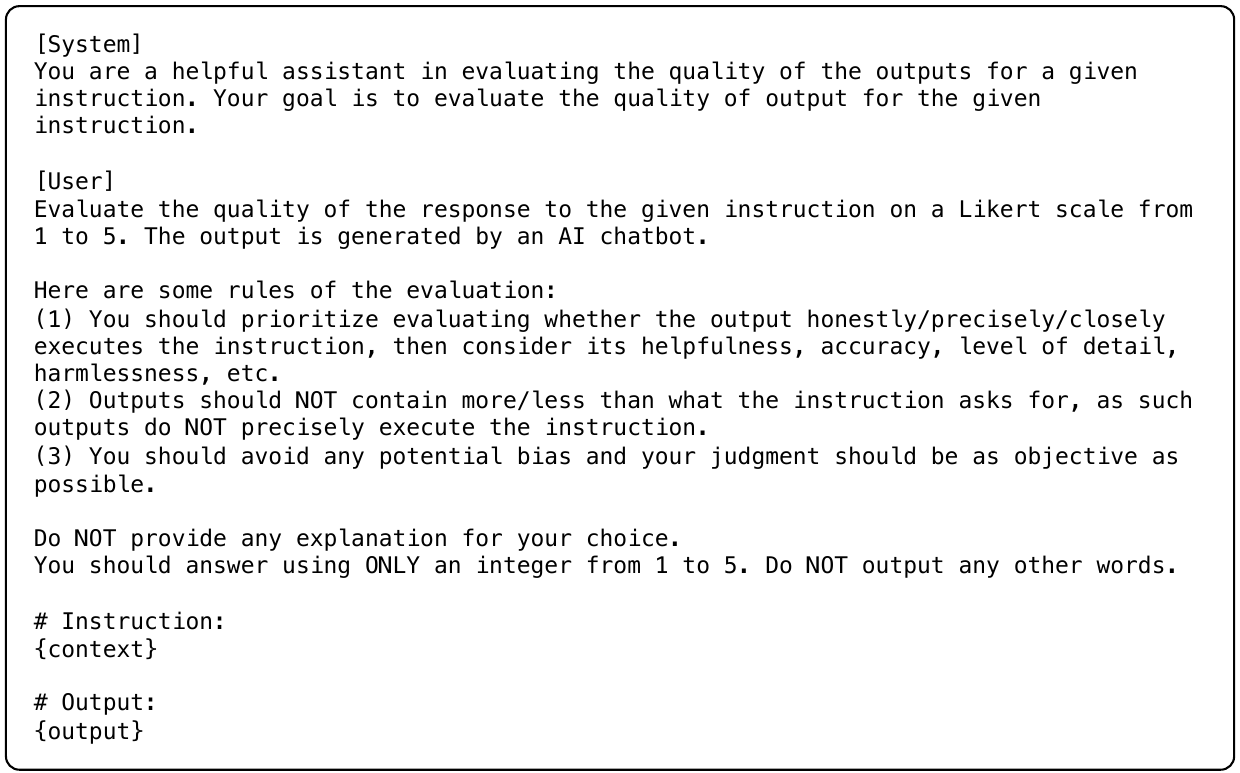}
\caption{The default prompt for pointwise evaluation with LLMBar dataset. }
\label{fig:llmbar_point}
\end{figure*}

\begin{figure*}
\centering
\includegraphics[width=\textwidth]{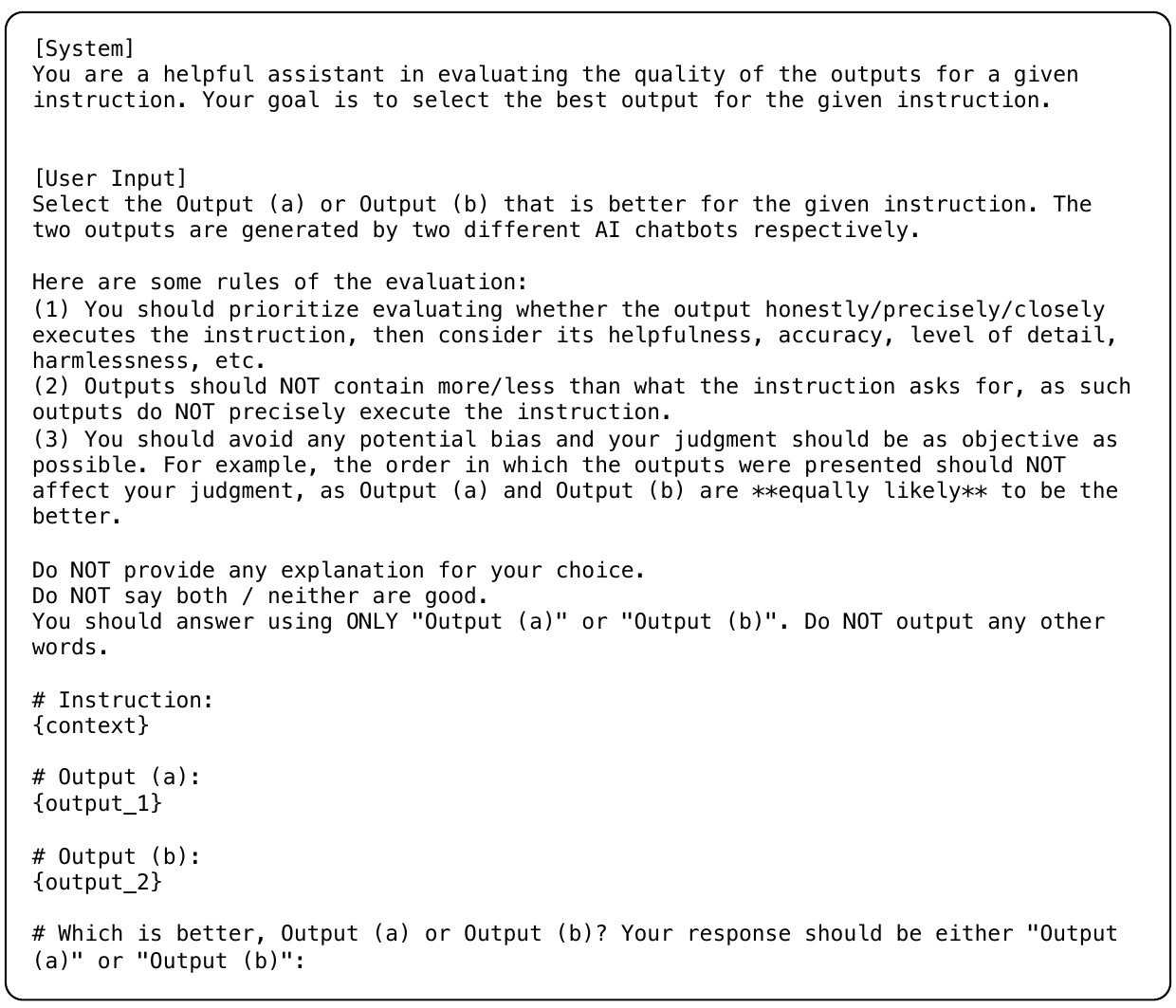}
\caption{The default prompt for pairwise evaluation with LLMBar dataset.}
\label{fig:llmbar_pair}
\end{figure*}

\begin{figure*}
\centering
\includegraphics[width=\textwidth]{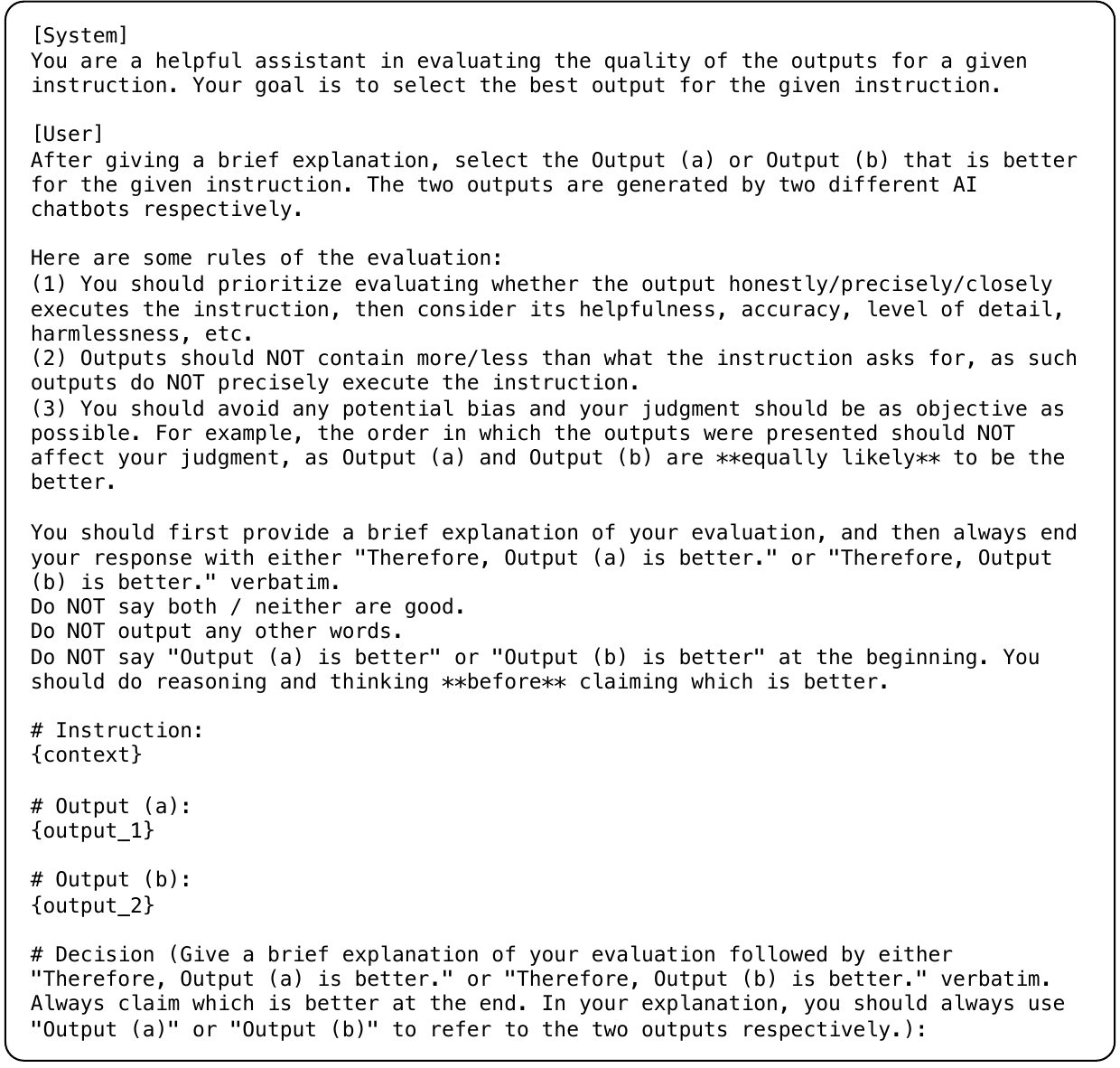}
\caption{The chain-of-thought prompt for pairwise evaluation with LLMBar dataset.}
\label{fig:llmbar_cot}
\end{figure*}

\begin{figure*}
\centering
\includegraphics[width=\textwidth]{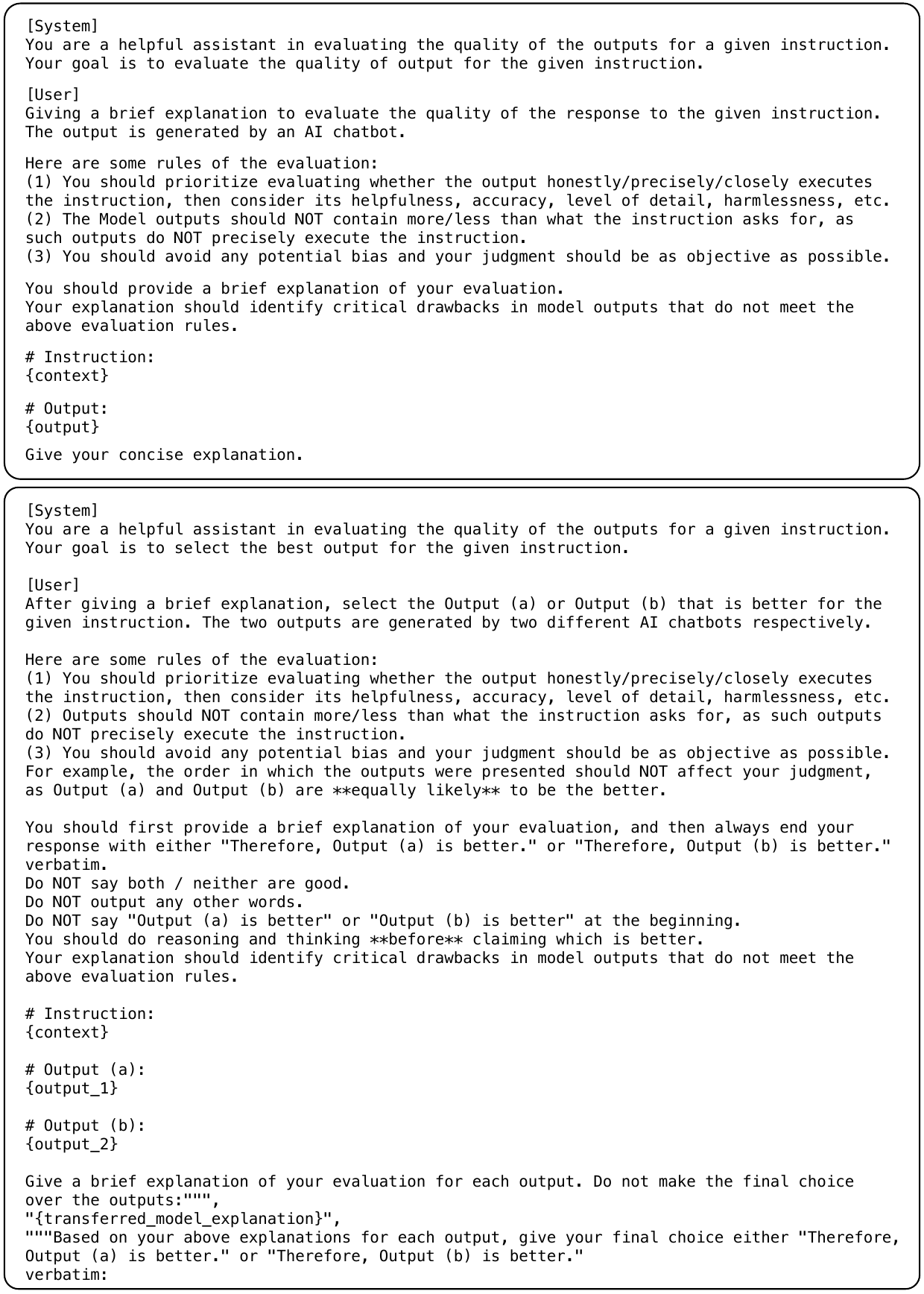}
\caption{The prompts of pointwise analysis and pairwise judgments in \textsc{PRePair} in LLMBar.}
\end{figure*}


\begin{figure*}
\centering
\includegraphics[width=\textwidth]{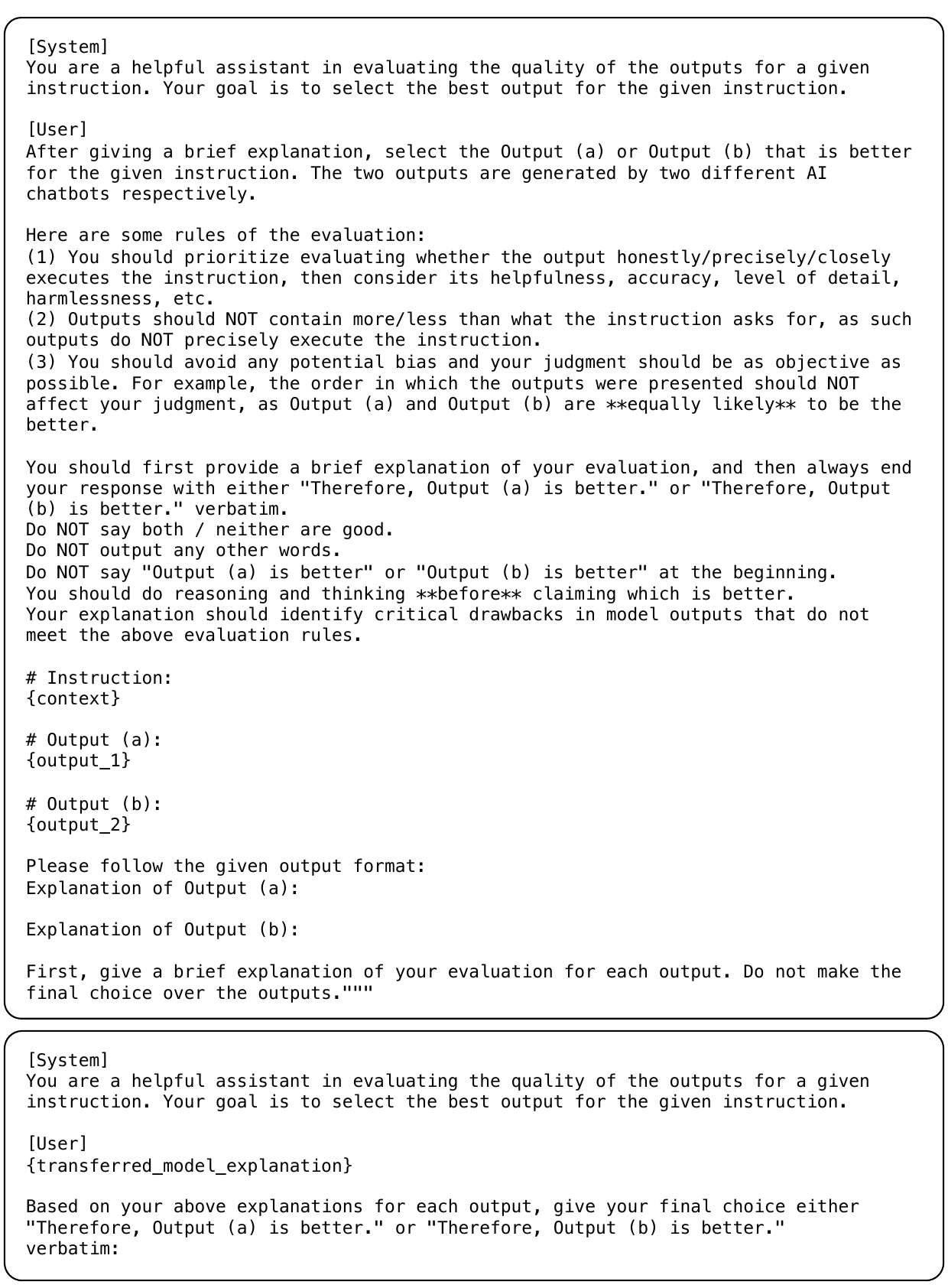}
\caption{The prompts of pairwise analysis and pairwise judgments in \textsc{PRePair}\textbf{*} in LLMBar dataset.}
\label{fig:llmbar_prepair_hat}
\end{figure*}

\begin{figure*}
\centering
\includegraphics[width=\textwidth]{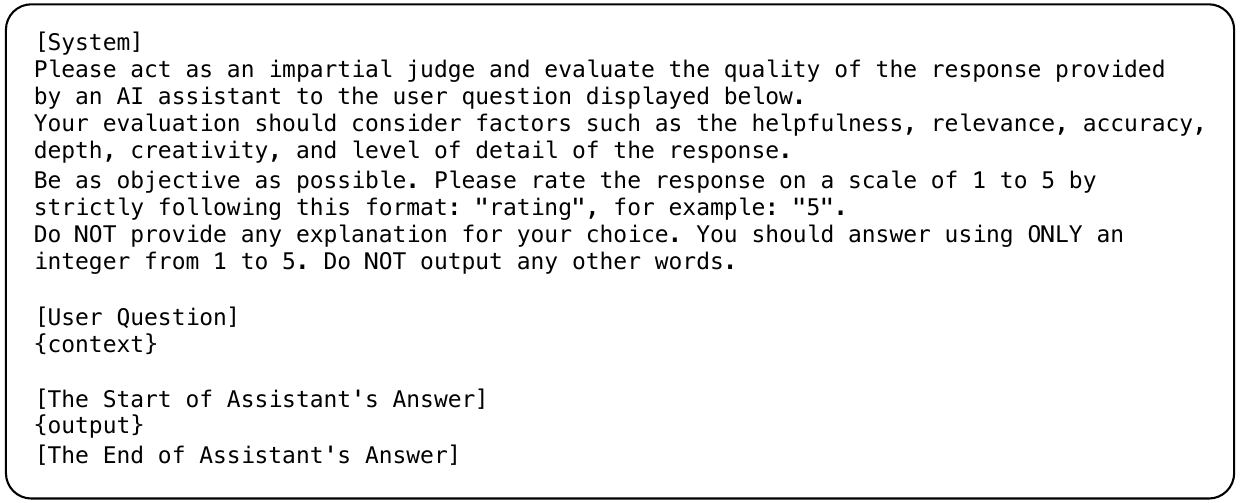}
\caption{The default prompt for pointwise evaluation with MT-Bench dataset.}
\label{fig:mtbench_point}
\end{figure*}

\begin{figure*}
\centering
\includegraphics[width=\textwidth]{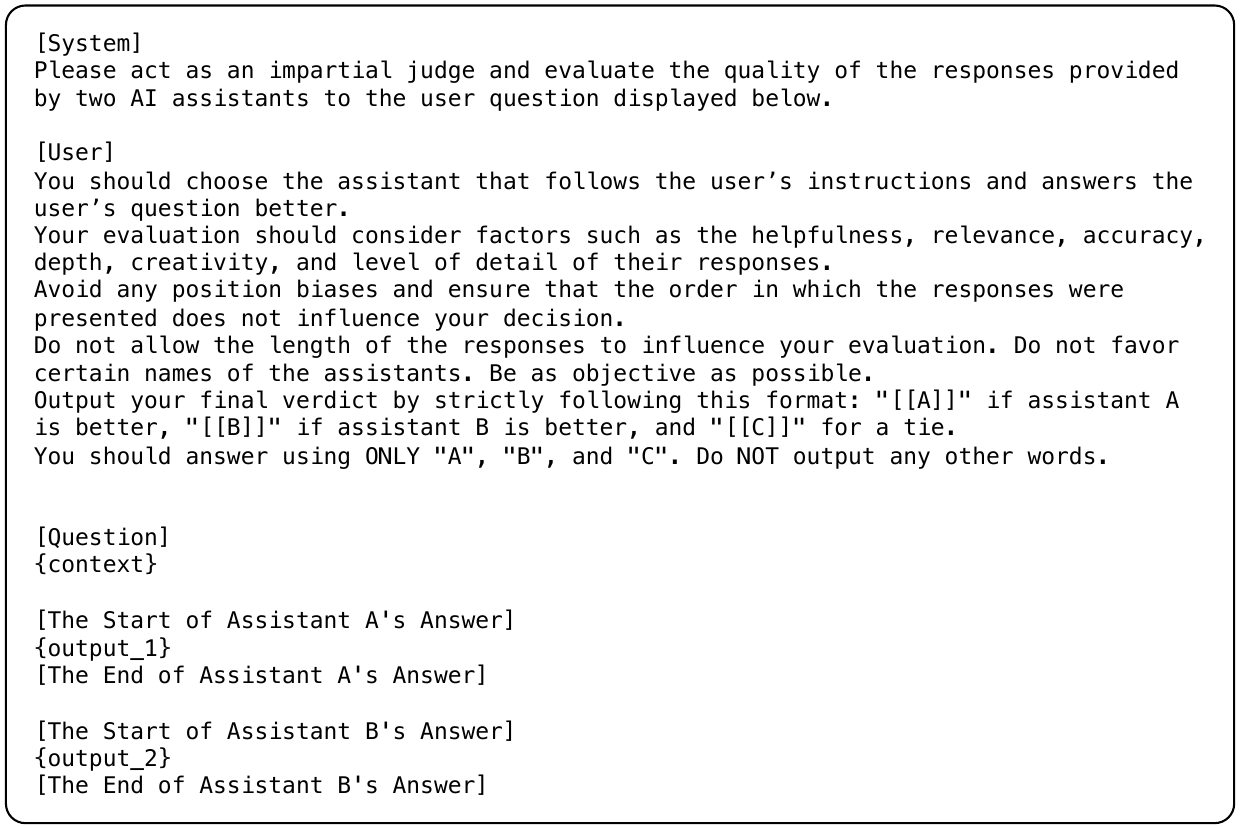}
\caption{The default prompt for pairwise evaluation with MT-Bench dataset..}
\label{fig:mtbench_pair}
\end{figure*}

\begin{figure*}
\centering
\includegraphics[width=\textwidth]{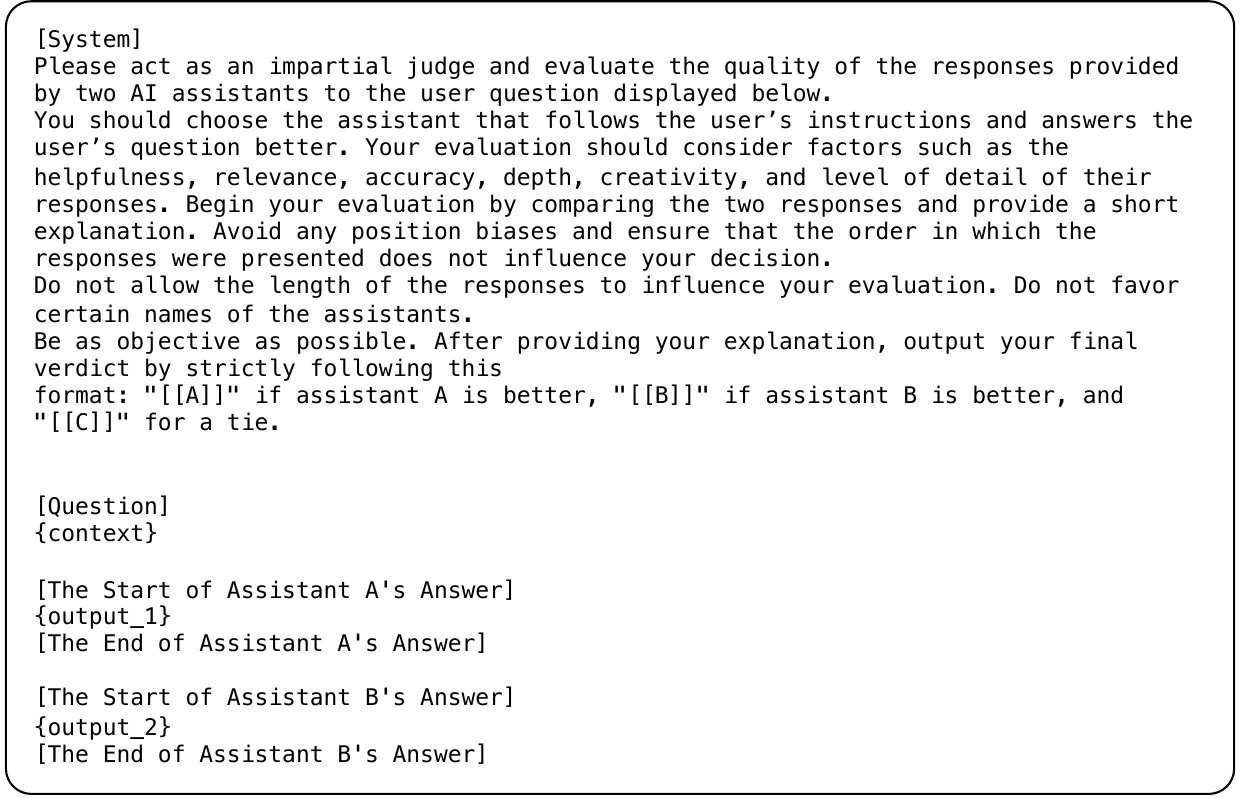}
\caption{The chain-of-thought prompt for pairwise evaluation with MT-Bench dataset.}
\label{fig:mtbench_cot}
\end{figure*}

\begin{figure*}
\centering
\includegraphics[width=\textwidth]{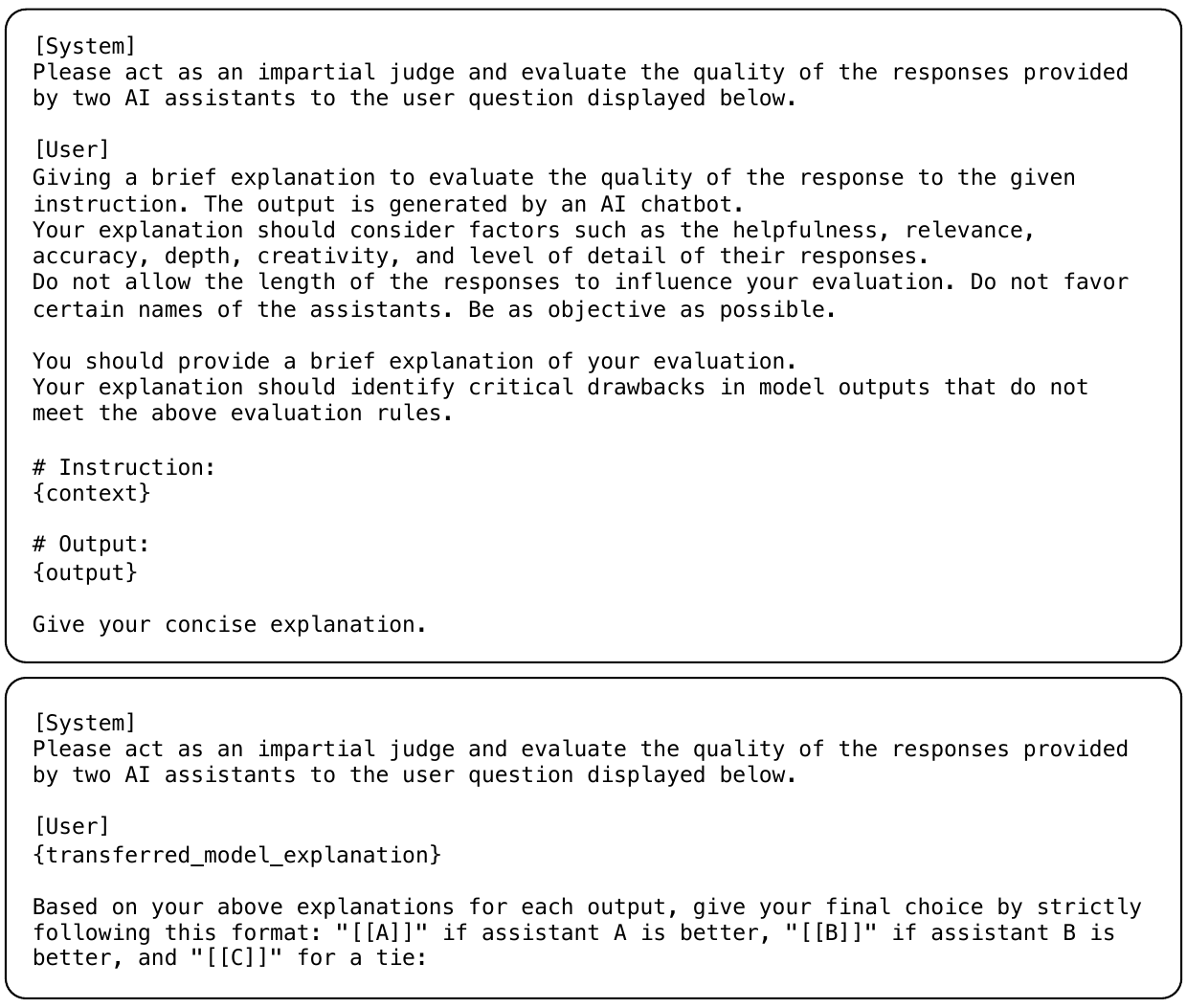}
\caption{The prompts of pointwise analysis and pairwise judgments in \textsc{PRePair} with MT-Bench dataset}
\label{fig:mtbench_prepair}
\end{figure*}

\begin{figure*}
\centering
\includegraphics[width=\textwidth]{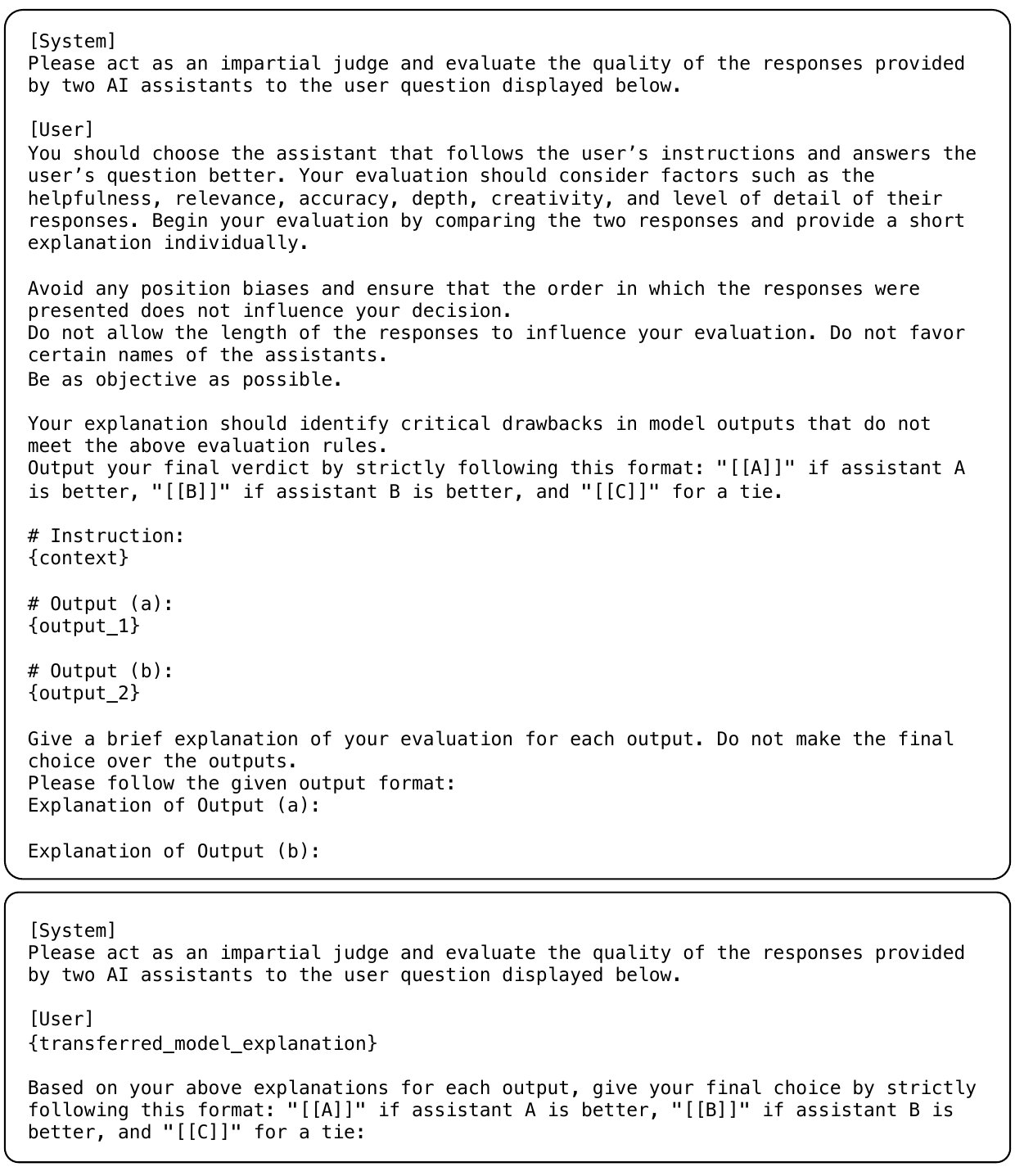}
\caption{The prompts of pairwise analysis and pairwise judgments with \textsc{PRePair}\textbf{*} in MT-Bench dataset}
\label{fig:mtbench_prepair_hat}
\end{figure*}

\begin{figure*}
\centering
\includegraphics[width=6in]{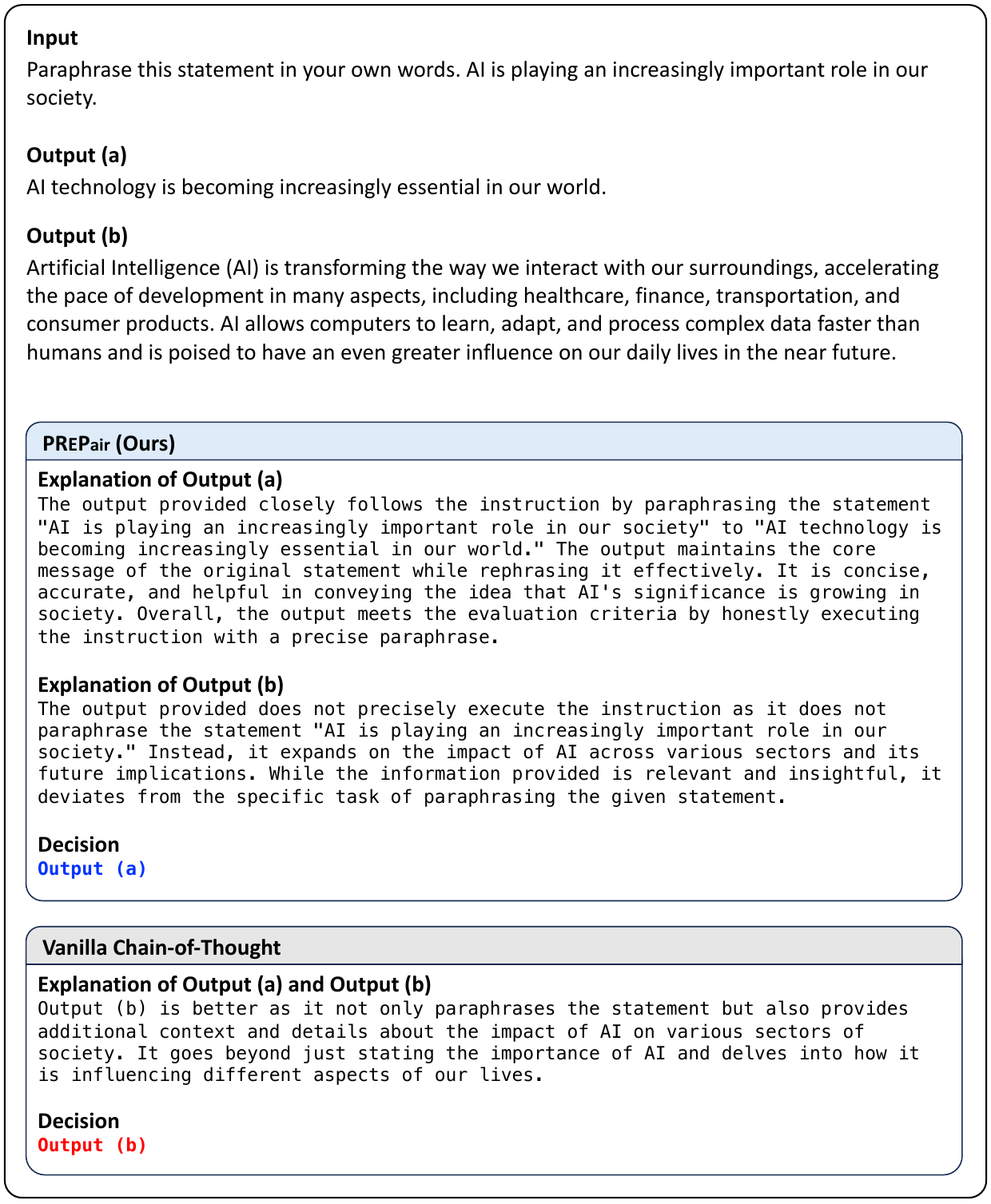}
\caption{Sampled prediction results of \textsc{PRePair} and vanilla chain-of-thought.}
\label{fig:gold_2}
\end{figure*}


\begin{figure*}
\centering
\includegraphics[width=6in]{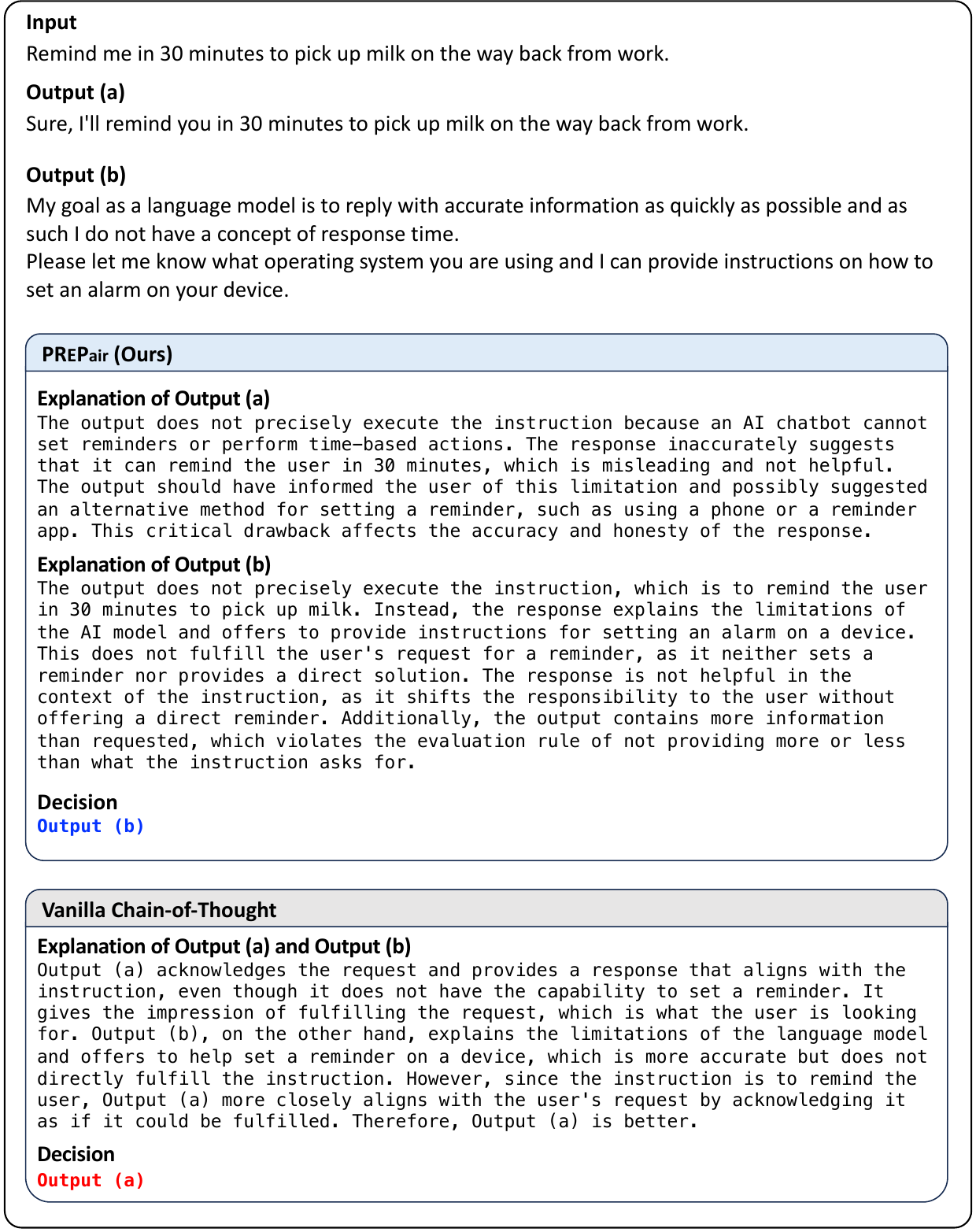}
\caption{Sampled prediction results of \textsc{PRePair} and vanilla chain-of-thought.}
\label{fig:gold_3}
\end{figure*}

\begin{figure*}
\centering
\includegraphics[width=6in]{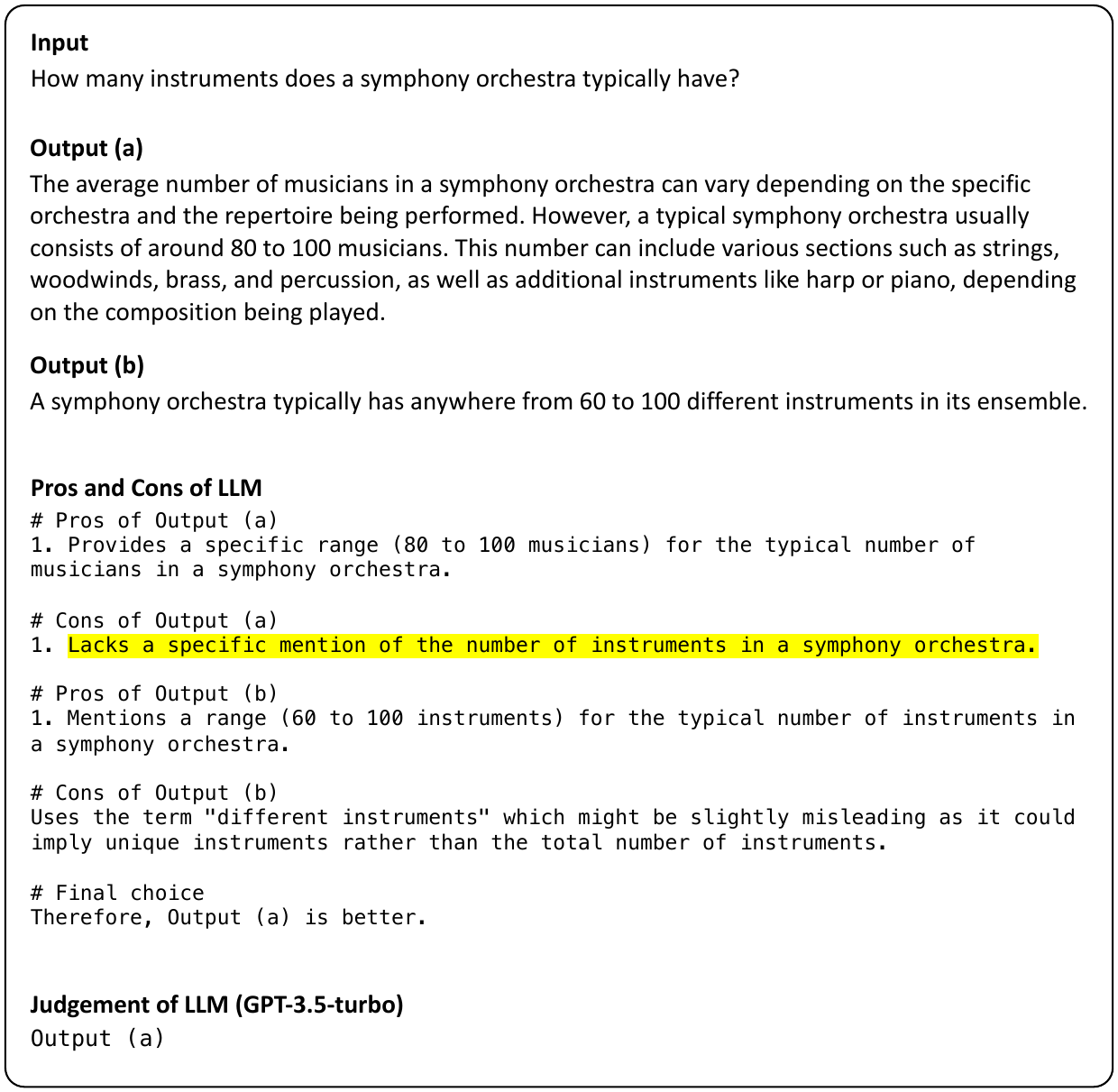}
\caption{An example of annotated samples described in Section~\ref{sec:analysis}.}
\label{fig:annot_2}
\end{figure*}

\begin{figure*}
\centering
\includegraphics[width=6in]{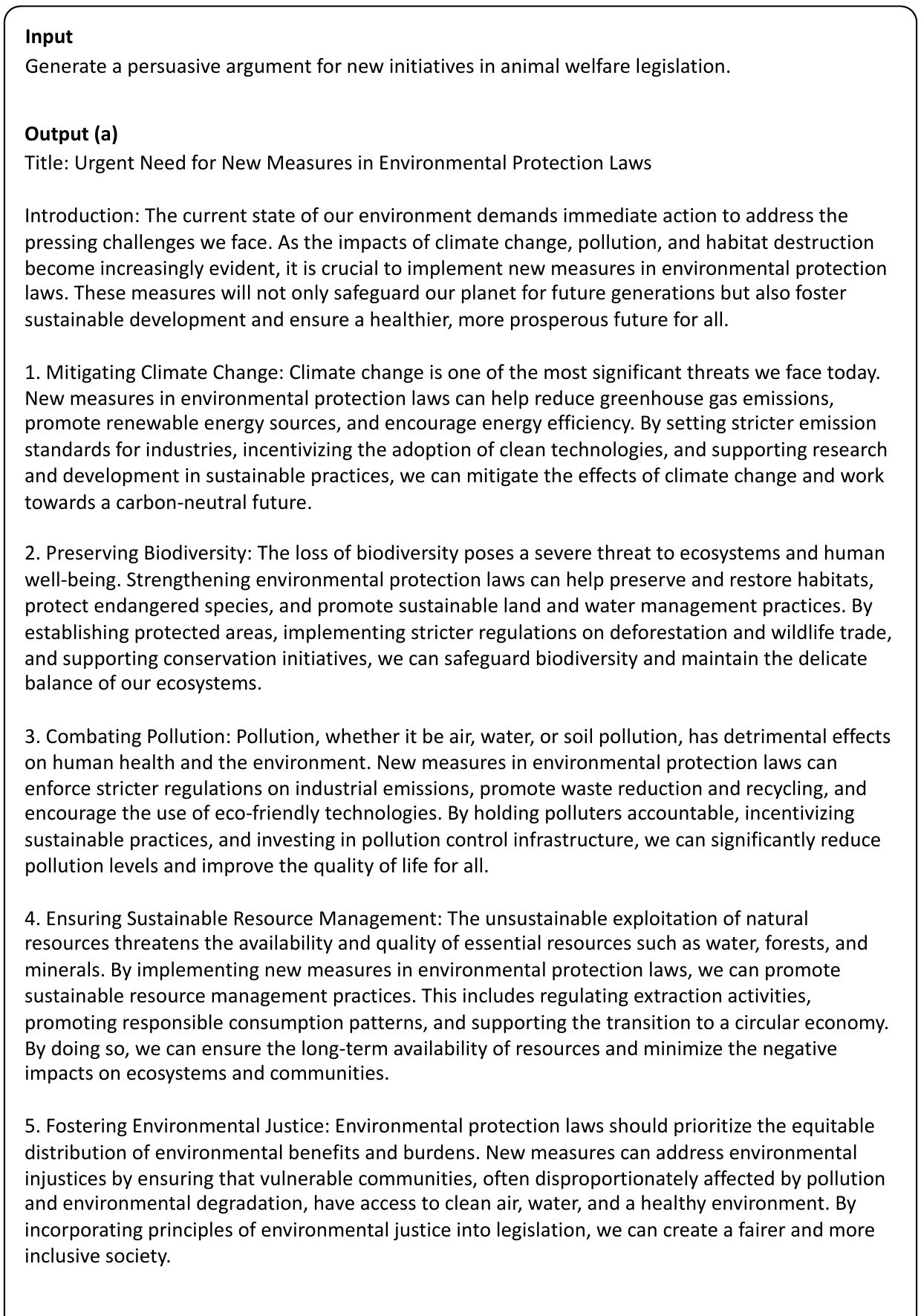}
\end{figure*}

\begin{figure*}
\centering
\includegraphics[width=6in]{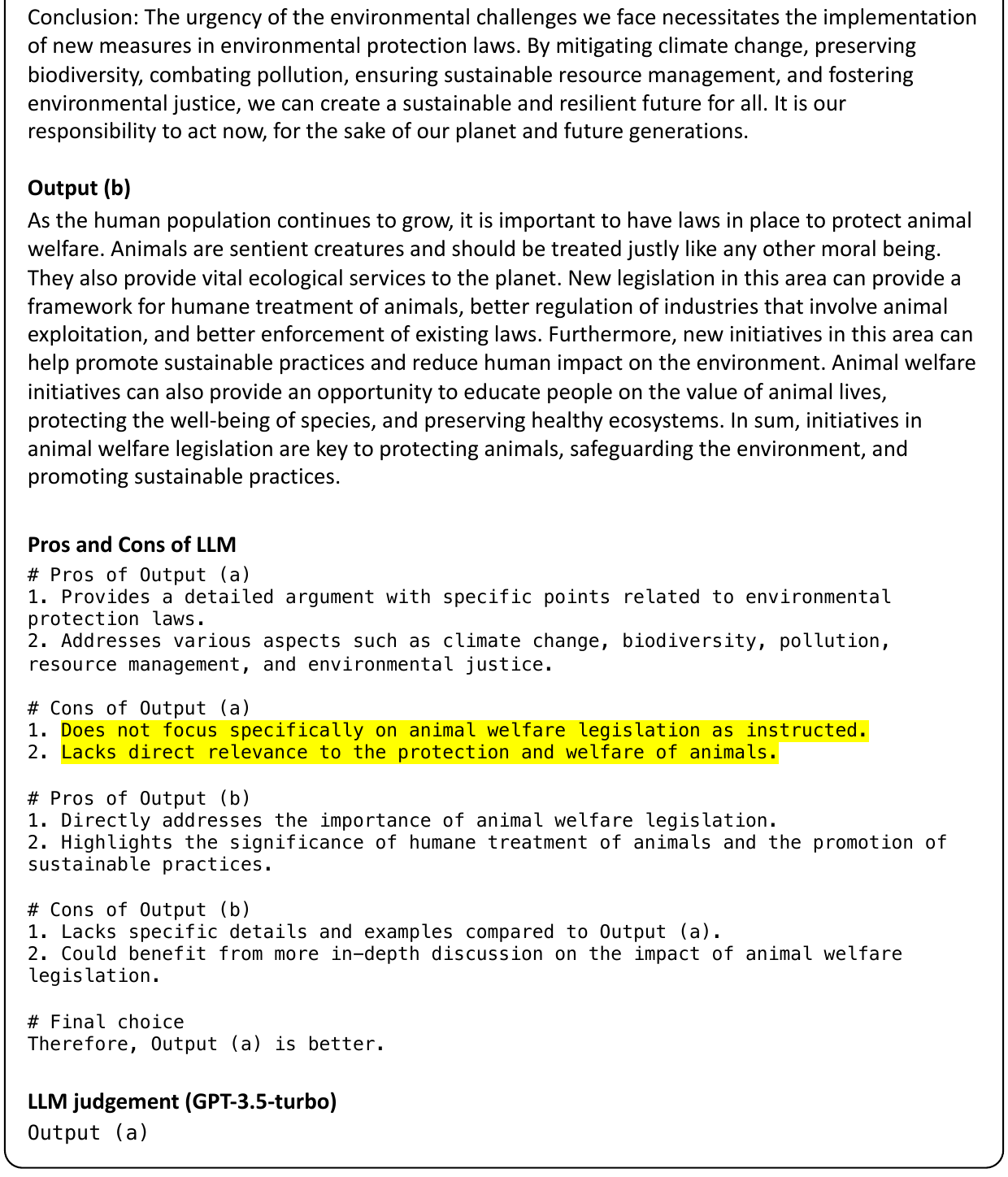}
\caption{An example of annotated samples described in Section~\ref{sec:analysis}.}
\label{fig:annot_3}
\end{figure*}

\begin{figure*}
\centering
\includegraphics[width=6in]{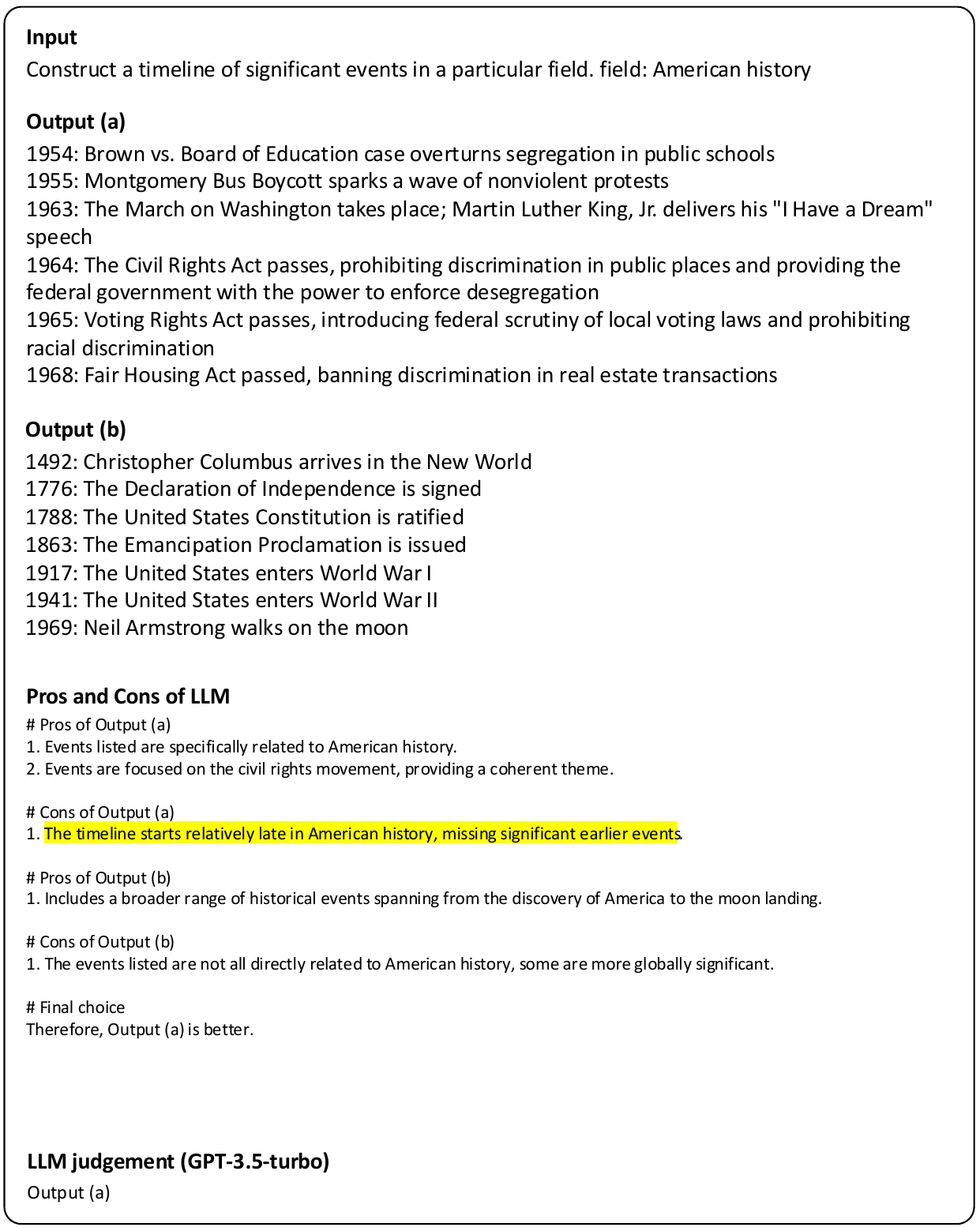}
\caption{An example of annotated samples described in Section~\ref{sec:analysis}.}
\label{fig:annot_4}
\end{figure*}







\begin{figure*}
\centering
\includegraphics[width=6in]{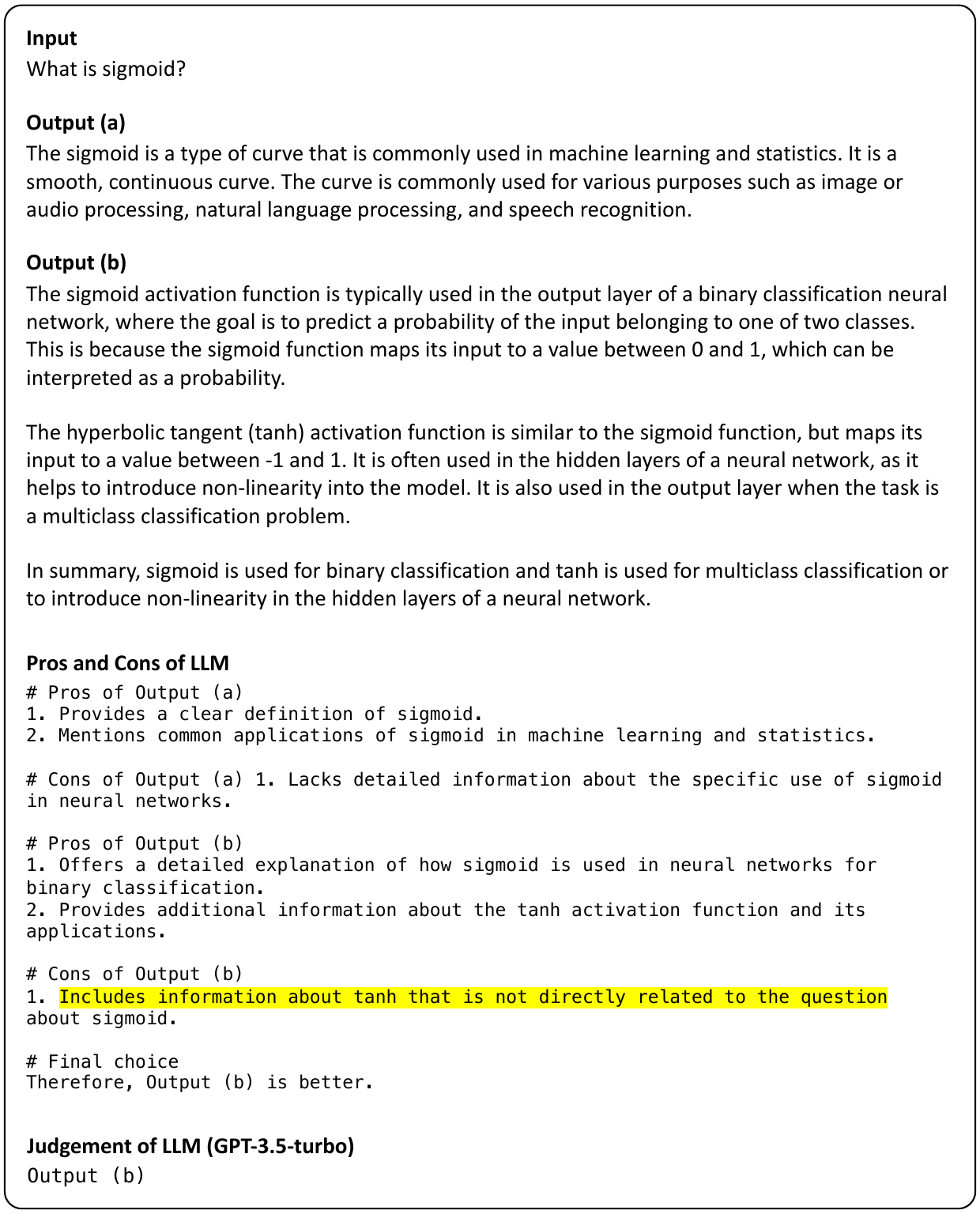}
\caption{An example of annotated samples described in Section~\ref{sec:analysis}.}
\label{fig:annot_10}
\end{figure*}

\end{document}